\documentclass[journal]{IEEEtran}

\ifCLASSINFOpdf
\else
\fi

\usepackage{booktabs}
\usepackage{amsmath}
\usepackage{rotating}
\usepackage[pagebackref=true,breaklinks=true,
letterpaper=true,colorlinks,citecolor=blue,
linkcolor=blue,bookmarks=false]{hyperref}

\usepackage{color}
\usepackage{times}
\usepackage{epsfig}
\usepackage{graphicx}
\usepackage{amssymb}
\usepackage{multirow}
\usepackage{algorithm}
\usepackage{algorithmic}
\usepackage{array}
\usepackage{cite}
\newcommand{\RN}[1]{%
  \textup{\uppercase\expandafter{\romannumeral#1}}%
}

\usepackage[utf8]{inputenc}
\usepackage[T1]{fontenc}
\usepackage{textcomp}
\usepackage[table]{xcolor}
\usepackage{threeparttable}

\begin{document}

\title{Structured Context Enhancement Network for Mouse Pose Estimation}
\author{Feixiang Zhou, Zheheng Jiang, Zhihua Liu, Fang Chen, Long Chen, Lei Tong, Zhile Yang, Haikuan Wang, Minrui Fei, Ling Li and Huiyu Zhou 
\thanks{F. Zhou, Z. Liu, F. Chen, L. Chen, L. Tong and H. Zhou are with School of Informatics, University of Leicester, United Kingdom. H. Zhou is the corresponding author. E-mail: hz143@leicester.ac.uk. }
\thanks{Z. Jiang is with School of Computing and Communications, Lancaster University, United Kingdom.}
\thanks{Z. Yang is with Shenzhen Institute of Advanced Technology, Chinese Academy of Sciences, Shenzhen, China.}
\thanks{H. Wang and M. Fei are with School of Mechanical Engineering and Automation, Shanghai University, Shanghai, China.}
\thanks{L. Li is with School of Computing, University of Kent, United Kingdom.}}


\maketitle
\begin{abstract}
Automated analysis of mouse behaviours is crucial for many applications in neuroscience. However, quantifying mouse behaviours from videos or images remains a challenging problem, where pose estimation plays an important role in describing mouse behaviours. Although deep learning based methods have made promising advances in human pose estimation, they cannot be directly applied to pose estimation of mice due to different physiological natures. Particularly, since mouse body is highly deformable, it is a challenge to accurately locate different keypoints on the mouse body. In this paper, we propose a novel Hourglass network based model, namely Graphical Model based Structured Context Enhancement Network (GM-SCENet) where two effective modules, i.e., Structured Context Mixer (SCM) and Cascaded Multi-level Supervision (CMLS) are subsequently implemented. SCM can adaptively learn and enhance the proposed structured context information of each mouse part by a novel graphical model that takes into account the motion difference between body parts. Then, the CMLS module is designed to jointly train the proposed SCM and the Hourglass network by generating multi-level information, increasing the robustness of the whole network. Using the multi-level prediction information from SCM and CMLS, we develop an inference method to ensure the accuracy of the localisation results. Finally, we evaluate our proposed approach against several baselines on our Parkinson’s Disease Mouse Behaviour (PDMB) and the standard DeepLabCut Mouse Pose datasets. The experimental results show that our method achieves better or competitive performance against the other state-of-the-art approaches.
\end{abstract}


\IEEEpeerreviewmaketitle

\section{Introduction}
\IEEEPARstart{M}{O}{U}{S}{E} models are an important tool to study neurodegenerative disorders such as Alzheimer \cite{Lewejohann2009} and Parkinson’s disease \cite{Blume2009} in neuroscience. Modelling mouse behaviour is key to understand brain functions and disease mechanisms \cite{Abbott2013}, due to certain similarity and homology between humans and animals. Historically, studying mouse behaviour can be a time-consuming and difficult task because the collected data requires experts to physically engage. Recent advances in computer vision and machine learning have facilitated automated analysis for complex behaviours \cite{jiang2018context,Nguyen2019,Robie2017}. This makes it possible to conduct a wide range of behavioural experiments without human intervention, which has yielded new insights into the pathologies and treatment of specific diseases  \cite{Schaefer2012,Kilpelainen2019,Egnor2016}.

In this paper, we are particularly interested in 2D mouse pose estimation, which is one of the fundamental problems in mouse behaviour analysis. Mouse pose estimation is defined as the problem of measuring mouse posture which denotes the geometrical configuration of body parts in images or videos \cite{mathis2020deep}. This provides useful information for ethologically relevant behaviours where the keypoint localisation of individual body parts is significant \cite{mathis2020deep}. Upon its success, pose analysis can be used to reveal mouse motor functions by providing configuring features \cite{arac2019deepbehavior}, especially for studying social behaviour in experimental mice. Specifically, the spatio-temporal motion representations derived from the keypoints of each mouse could be encoded to interpret social behaviours. Mouse pose estimation is non-trivial at all due to small size, occlusions, invisible keypoints, and deformation of the body. Early approaches focused on labelling the locations of interest with unique and recognizable markers in order to simplify the pose analysis \cite{Maghsoudi2018, Ohayon2013,KARAKOSTAS201431}. However, these systems are 'invasive' to the targeted subjects. To overcome this problem, several advanced markerless methods have been proposed for mouse pose estimation \cite{Burgos-artizzu2012,DeChaumont2012,Mathis2018 }. Overall, a common technique used in these methods is to fit specific skeleton or active models, such as bounding box \cite{Burgos-artizzu2012}, ellipse \cite{DeChaumont2012} and contour \cite{Branson2005}. These methods have produced promising results, but the flexibility of such methods is limited as they require sophisticated skeleton models.

With the development of Deep Convolutional Neural Networks (DCNNs), significant progresses have been achieved in human and animal pose estimation \cite{Cao2019,Liu2020,Zhang2020,Graving2019}. Most of the existing systems (e.g., DeepLabCut \cite{Mathis2018}, LEAP \cite{Pereira2019}, DeepPoseKit \cite{Graving2019}) are built on deep networks for human pose estimation. For example, DeepLabCut combines the feature detectors of DeeperCut \cite{insafutdinov2016deepercut} with readout layers, namely deconvolutional layers, for markerless pose estimation, and exploits transfer learning to train the deep model with fewer examples. Similarly, LEAP has been also developed based on an existing network, namely SegNet \cite{Badrinarayanan2017}, which provides a graphical interface for labelling body parts. However, its preprocessing is computationally expensive, thus limiting the application of their system in other environments. DeepPoseKit uses two Hourglass modules with different scales to learn multi-scale geometry of the keypoints, based on the standard Hourglass network \cite{newell2016stacked}.

Although the system performance on different animal datasets reported in these works is encouraging, the set-up of their environments is simple with clean background. These established systems mainly deployed the standard human pose estimation networks for animal applications. A common pipeline of the existing deep models is to generate heatmaps (also known as score maps) of all the generated keypoints simultaneously at the final stage of the networks \cite{Graving2019,Mathis2018, Pereira2019, newell2016stacked, Wei2016,Sun2019}. However, these systems have not fully considered the differences between different body parts, which is critical to spatio-temporal analysis. This gap may be due to the fact that spatial correlations between mouse body parts are much weaker than those of human body parts due to highly deformable body structures.

\begin{figure*}
\begin{center}
\includegraphics[width=15cm]{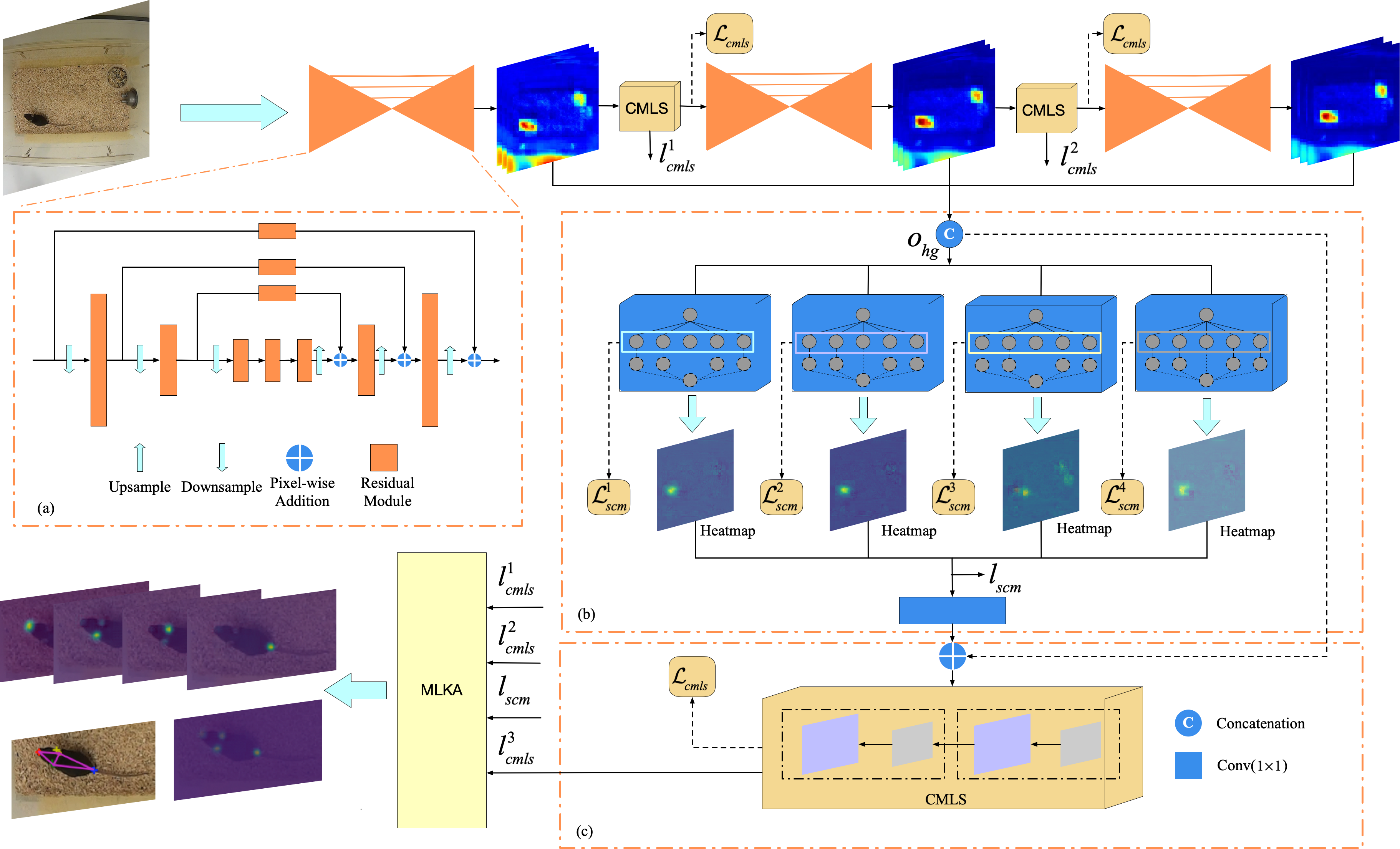}
\end{center}
\caption{Overview of the proposed framework for mouse pose estimation. (a) The structure of the Hourglass network \cite{newell2016stacked}. We stack 3 Hourglass modules to capture multi-level pose features. (b) The structure of the proposed Structured Context Mixer (SCM). It first generates global contextual information by fusing multi-level features from (a). To describe the differences between different body parts, we then design 4 independent subnetworks to predict the heatmap of the corresponding keypoint, where each subnetwork explores global and keypoint-specific contextual information to represent structured contextual information of each keypoint (see Section \ref{section3.1.1}). The blue blocks refer to the 4 subnetworks, in each of which we propose a novel graphical model to adaptively fuse the global and keypoint-specific contextual information (see Section \ref{section3.1.2}). (c) The structure of Cascaded Multi-level Supervision (CMLS). We use this module as an intermediate supervision agent with the multi-scale outputs for joint training of SCM, the CMLS module and the Hourglass network (see Section \ref{section3.2}). We attach this module to the end of each Hourglass module and the proposed SCM. 
Moreover, during the inference phase, the prediction results (i.e. locations of keypoints) $l_{cmls}^1$, $l_{cmls}^2$ and  $l_{cmls}^3$ generated in the CMLS modules and ${l_{scm}}$ generated in SCM are used as input of our proposed  Multi-level Keypoint Aggregation (MLKA) inference method to produce the final localisation results (see Section \ref{section3.3}). All these components are trained in an end-to-end fashion. ${{\cal L}_{cmls}}$ in (c) represents the loss of all the keypoints. ${\cal L}_{scm}^1$, ${\cal L}_{scm}^2$, ${\cal L}_{scm}^3$  and ${\cal L}_{scm}^4$ in (b) represent the losses of keypoints 1-4, respectively.}

\label{fig:framework}
\end{figure*}

 We here propose a Graphical Model based Structured Context Enhancement Network (GM-SCENet) for mouse pose estimation, as shown in Fig. \ref{fig:framework}. We use the standard Hourglass network \cite{newell2016stacked} as the basic structure due to its strong multi-scale feature representation capability (Fig. \ref{fig:framework}(a)). In our approach, the proposed Structured Context Mixer (SCM), shown in Fig. \ref{fig:framework}(b), first captures global spatial configurations of the full mouse body to describe the global contextual information by concatenating multi-level feature maps from the Hourglass network. In combination with the global contextual information, our proposed SCM explores the neighbouring areas at four different directions of each keypoint collected from mouse body parts. Afterwards, we model the relationship between the keypoint-specific and the global contextual information by a novel graphical model. We define these two types of contextual information as structured contextual information hereafter.

In addition, we introduce a novel Cascaded Multi-level Supervision module (CMLS) \cite{Wei2016,newell2016stacked} to jointly train the proposed SCM and the Hourglass network, as shown in Fig. \ref{fig:framework}(c). This allows our network to generate multi-level information for addressing the problems of scale variations and small subjects by strengthening contextual learning. In fact, the prediction results of the current scale/level, as prior knowledge, are used for semantic description with the supervision of high-scale feature maps. In the meantime, the extracted features and the predicted heatmaps within the CMLS module can be further refined using a cascaded structure. Finally, to integrate multi-level localisation results, we also design an inference algorithm called Multi-level Keypoint Aggregation (MLKA) to produce the final prediction results, where the locations of the selected multi-level keypoints from all the supervision layers are aggregated.

Fig. \ref{fig:framework} shows our proposed framework for mouse pose estimation. The main contributions of our work can be summarised as follows:
\begin{itemize}
\item We propose a novel Graphical Model based Structured Context Enhancement Network (GM-SCENet) which consists of Structured Context Mixer (SCM), a Cascaded Multi-level Supervision module (CMLS) and the backbone Hourglass network. The proposed GM-SCENet has the ability of modelling the differences and spatial correlations between mouse body parts simultaneously.

\item The proposed SCM can adaptively learn and enhance the structured contextual information of each keypoint. This is achieved by exploring the global and keypoint-specific contextual information to describe the differences between body parts, and designing a novel graphical model to explicitly model the relationship between them.

\item We also develop a novel CMLS module to jointly train SCM and the Hourglass network, which adopts multi-scale features for supervised learning and increases the robustness of the whole network. In addition, we design a Multi-Level Keypoint Aggregation (MLKA) algorithm to integrate multi-level localisation results.
    
\item We introduce a new challenging mouse dataset for pose estimation and behaviour recognition, namely Parkinson’s Disease Mouse Behaviour (PDMB). Several common behaviours and the locations of four representative body parts 
are included in this dataset. To the best of our knowledge, this is the first large dataset for mouse pose estimation, collected from standard CCTV cameras.

\end{itemize}
\section{Related Work}
\subsection{Human Pose Estimation}
Many traditional methods for single-person pose estimation adopt graph structures, e.g., pictorial structures \cite{yang2011articulated} or graphical models \cite{chen2014articulated} to explicitly model the spatial relationships between body parts. However, the prediction results of keypoints rely heavily on hand-crafted features, which are susceptible to difficult problems such as occlusion. Recently, significant improvements have been achieved by designing DCNNs for learning high-level representations \cite{zhang2018poseflow,dong2017adore, liu2018human, Sun2019, chu2017multi,ke2018multi,song2017thin,artacho2020unipose}. DeepPose \cite{toshev2014deeppose} is the first attempt to use deep models for pose estimation, which directly regress the keypoints coordinates using an end-to-end network. However, generating predictions in this way misses the spatial relationship in the images. Recent methods mainly adopt fully convolutional neural networks (FCNNs) to regress 2D Gaussian heatmaps \cite{Wei2016,newell2016stacked}, followed by further inference using Gaussian. For instance, Newell et al. \cite{newell2016stacked} first proposed the Hourglass network to gradually capture representations across scales by repeating bottom-up and top-down processes. They also adopted intermediate supervision to train the entire network, and learnt the spatial relationship with a larger receptive field. In \cite{Sun2019}, Sun et al. designed the High-Resolution Net (HRNet) that pursues high-resolution representations, fusing multi-scale feature maps of parallel multi-resolution subnetworks. In spite of their promising performance, it remains an open problem to locate keypoints accurately due to abnormal postures, small objects and occluded body parts.

\subsection{Mouse Pose Estimation}
In previous works, fitting active shape models was a popular way to approach the understanding of mouse postures. For example, De Chaumont et al. \cite{DeChaumont2012} employed a set of geometrical primitives for mouse modelling. Mouse posture is represented by three parts such as head, trunk and neck with corresponding spatial distances between them. Similarly, in \cite{Branson2005}, 12 B-spline deformable 2D templates were adopted to represent a mouse contour which is mathematically described with 8 elliptical parameters. However, these methods require sophisticated skeleton models, thus limiting the ability to fit to different scenes. Recently, most deep learning based methods for 2D mouse or other animal pose estimation are based on  human pose estimation models such as DeeperCut \cite{insafutdinov2016deepercut}, SegNet \cite{Badrinarayanan2017}, and Hourglass network \cite{newell2016stacked}. In \cite{Graving2019}, Graving et al. presented a multi-scale deep learning model called Stacked DenseNet to generate confidence maps from keypoints, based on the standard Hourglass architecture. This network is able to capture multi-scale information with intermediate supervision capabilities. Nevertheless, the ability of these approaches to deal with the problems such as occlusion and small target size is still weak.


\section{Proposed Methods}
We formulate the task of mouse pose estimation in images as the problem of learning a non-linear mapping  ${\cal F}:{\cal I} \to {\cal G}$, where ${\cal I}$ and ${\cal G}$ refer to image and keypoint spaces.  More formally, let $T = \{ ({I_i},{\tilde G_i})\} _{i = 1}^M$ be the training set of $M$ mouse images, where  ${I_i} \in {\cal I}$ denotes the input image and ${\tilde G_i} \in {\cal G}$ represents the corresponding ground-truth labels. ${\tilde G_i}$ can be represented as $\tilde{G}_{i}=\left\{G_{i}^{1}, G_{i}^{2}, \ldots, G_{i}^{K}\right\} \in \mathbb{R}^{K \times 2}$, where $K$ is the number of the keypoints defined in the image space ${\cal I}$.

To learn ${\cal F}$, we propose a framework composed of three main parts: Structured Context Mixer, Cascaded Multi-Level Supervision module and the backbone network. The proposed SCM aims to model the dependency between keypoint-specific and global contextual information, which focuses on both spatial correlations and differences between keypoints. In this paper, we denote $O = \{ {O_c}\} ,c \in C = \{ {c_g},{c_t},{c_l},{c_b},{c_r}\}$ as the structured contextual feature maps of each keypoint, where ${c_g}$ represents the global contextual information, ${c_t},{c_l},{c_b}$ and ${c_r}$ are keypoint-specific contextual information. Here, ${O_c}$ can be represented as a set of feature vectors ${O_c} = \{ o_c^i\} _{i = 1}^N$, $o_{c}^{i} \in \mathbb{R}^{L}$, where $N$ is the number of the image pixels in a contextual feature map and $L$ is the number of the channels of the contextual feature maps. Unlike previous systems that model contextual information using direct concatenation \cite{gidaris2015object}, channel attention \cite{hu2018squeeze} and spatial attention \cite{chu2017multi}, we here propose to adaptively fuse the structured contextual information by learning a set of latent feature maps ${H_c} = \{ h_c^i\} _{i = 1}^N,c \in C$ with a novel graphical model. In particular, we implement two interactive attention gates ${A_c} = \{ a_c^i\} _{i = 1}^N,a_c^i \in \{ 0,1\}$  and ${G_c} = \{ g_c^i\} _{i = 1}^N,g_c^i \in \{ 0,1\} $ to control the flow of the information between keypoint-specific and global contextual information. With the two-gate mechanism, our network can adaptively learn and enhance the relatively important keypoint-specific contextual features by preventing the potential loss of low- and high-level information. In SCM, the proposed graphical model jointly infers the hidden contextual features $H = \{ {H_c}\} ,c \in C$ and the attention gates $A = \{ {A_c}\} ,G = \{ {G_c}\}, c \in {C_k} = \{{c_t},{c_l},{c_b},{c_r}\}$.

\subsection{Structured Context Mixer}
\label{section3.1}

\subsubsection{Structured Context Information Representations}
\label{section3.1.1}
Extracting keypoints of mouse body parts is non-trivial at all due to occlusion, small size and highly deformable body. In order to accurately locate different keypoints of the targets, most deep models take multi-scale or high-resolution information into account \cite{Graving2019}, \cite{newell2016stacked}, \cite{Sun2019}, whilst utilising contextual information. In general, contextual information is referred to as regions surrounding the targets, and it has been proved effective in pose estimation \cite{newell2016stacked}, \cite{chu2017multi} and object detection\cite{9261609}. However, these models miss the differences between different keypoints to some extent, which may significantly contribute towards mouse pose estimation due to spatial correlations and restrictions. Unlike previous methods, we explore the structured contextual information of body parts to describe the differences between them, including global and keypoint-specific contextual information.

Before generating structured contextual information of each keypoint, we aggregate the low- and high-level feature maps from multiple stacks of the Hourglass network for representing the global contextual information ${O_{hg}}$ of the whole subject, as shown in Fig. \ref{fig:framework}. In fact, the aggregated features also encode the local contextual information derived from the low-level layers. ${O_{hg}}$ is defined as:
\begin{equation}
\begin{split}
{O_{hg}} = Concat({o_{s1}},{o_{s2}},...,{o_{sn}})
\label{equation:global context}
\end{split}
\end{equation}
where $Concat$ represents channel-wise concatenation, ${o_{s1}}$, ${o_{s2}}$, and ${o_{sn}}$ are the outputs of the first, second and $n$-th stack of the Hourglass network, respectively.

\begin{figure*}
\begin{center}
\includegraphics[width=13.6cm]{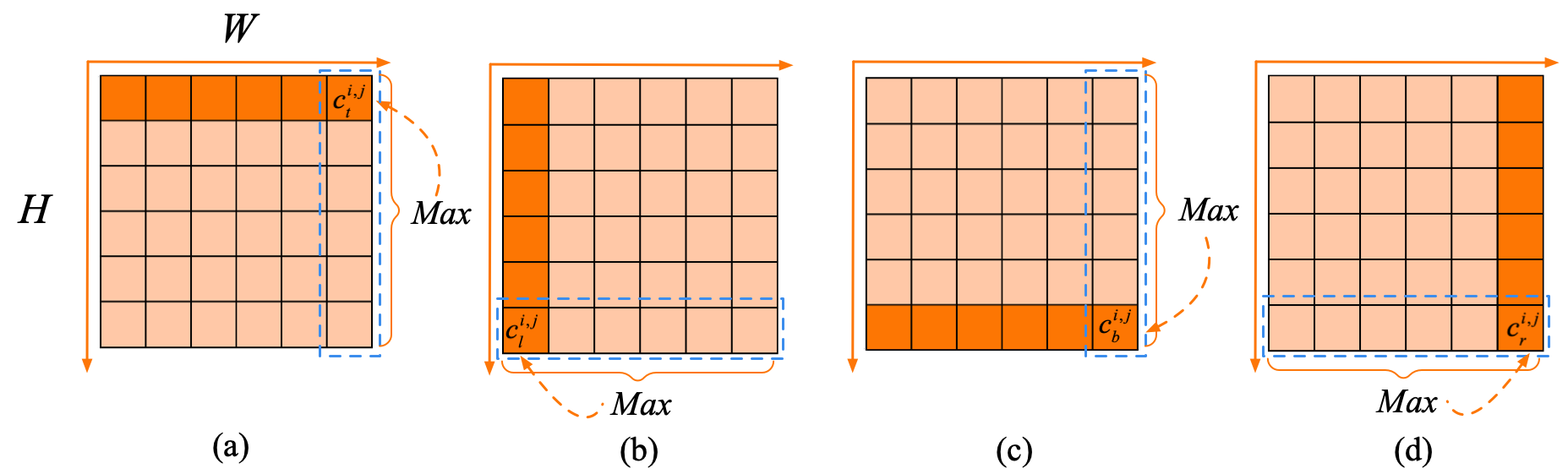}
\end{center}
\caption{Directionmax operation. (a) Topmax. (b) Leftmax. (c) Bottommax. (d) Rightmax. For Topmax and Leftmax, we scan from bottom to top for the vertical max-pooling and from right to left for the horizontal max-pooling respectively.}
\label{fig:maxdirection}
\end{figure*}

Afterwards, we extract keypoint-specific contextual information, where multiple discriminative regions are selected to represent the contextual information of each keypoint. This is beneficial to the accurate localisation of the keypoints. Inspired by the corner pooling scheme used in \cite{law2018cornernet}, we explore four contextual features to describe the keypoint-specific contextual information by the Directionmax operation, as shown in Fig. \ref{fig:maxdirection}, using prior knowledge.

In \cite{law2018cornernet}, to locate the top-left corner of a bounding box, we look horizontally towards the right for the topmost boundary of the target and vertically towards the bottom for the leftmost boundary. Similarly, we look horizontally towards the left for the bottommost boundary of the target and vertically towards the top for the rightmost boundary to pursue the bottom-right corner. Different from their work, we focus on locating different mouse keypoints in our paper. The problem of locating the top-left and bottom-right corners can be considered as that of locating a point when these two corners spatially overlap. Therefore, following the top-left and bottom-right pooling used in \cite{law2018cornernet}, we adopt top-left pooling (Topmax and Leftmax) and bottom-right pooling (Bottommax and Rightmax) simultaneously to generate four regions to indicate keypoint-specific contextual information. For a keypoint, Topmax means that we look vertically towards the bottom to explicitly investigate the area below the keypoint, while Leftmax suggests that we look horizontally towards the right to examine the area to the right of the keypoint. Similarly, Bottommax and Rightmax take us to look vertically towards the top and horizontally towards the left, respectively. With \(H \times W\) feature maps, the computation of Topmax shown in Fig. \ref{fig:maxdirection} uses the following equations:
\begin{equation}
\begin{split}
c_t^{(i,j)} = \left\{ {\begin{array}{*{20}{c}}
{Max\left( {{f_{c_t^{(i,j)}}},c_t^{(i + 1,j)}} \right)}&{{\rm{ if }} i < H}\\
{{f_{c_t^{(H,j)}}}}&{{\rm{ otherwise }}}
\end{array}} \right.
\label{equation:topmax}
\end{split}
\end{equation}

where ${f_{c_t^{}}}$ is the feature maps that are the inputs to the Topmax operations. ${{c_t^{}}}$ is the feature maps of the output, i.e., keypoint-specific contextual information. ${f_{c_t^{(i,j)}}}$ is the vectors at $(i,j)$.

\subsubsection{Structured Contextual Information Modelling with Graphical Model}
\label{section3.1.2}
\begin{figure*}
\begin{center}
\begin{tabular}{ccc}
\includegraphics[width=6.8cm]{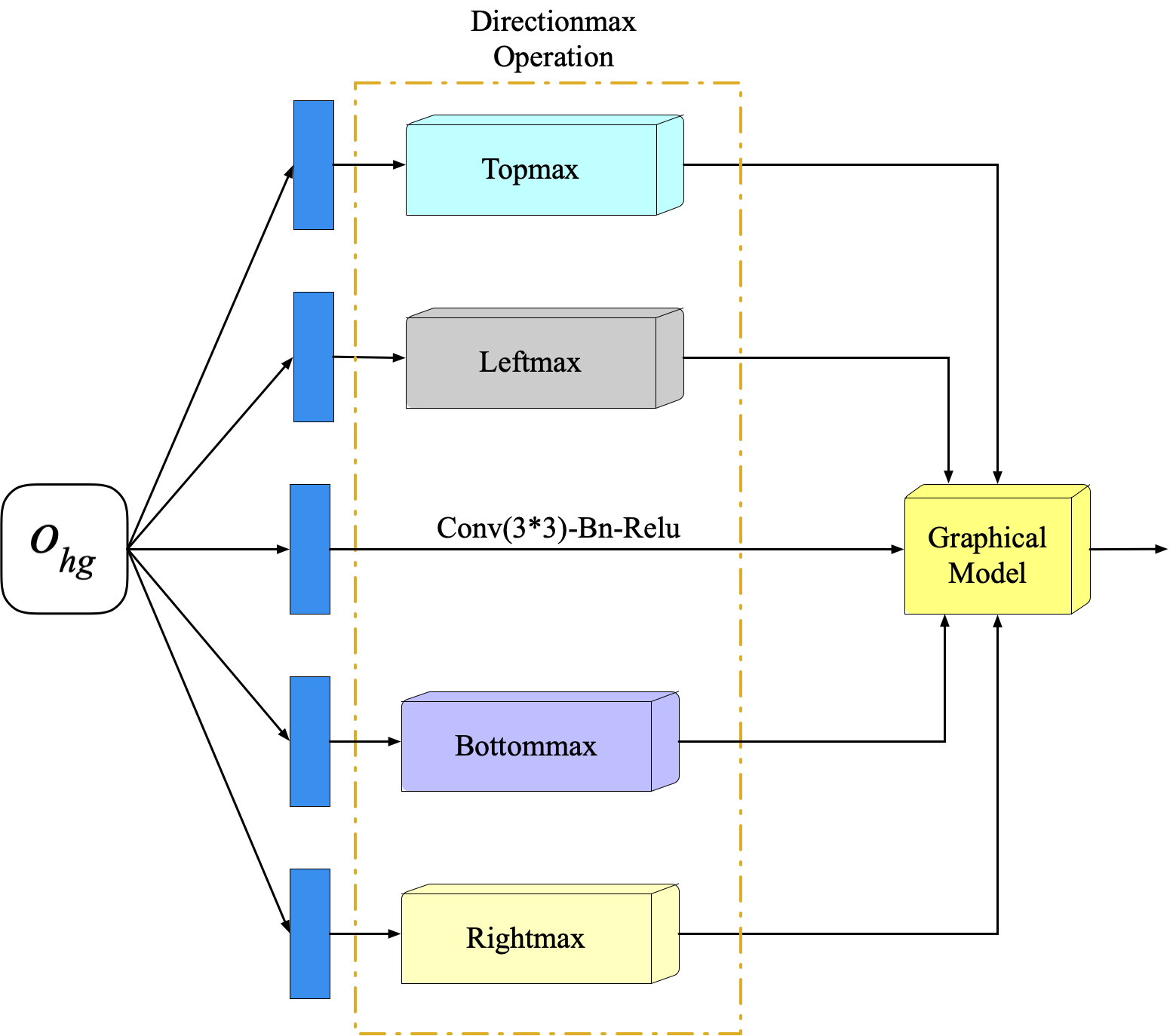}&
\hspace{6mm}
\includegraphics[width=6.8cm]{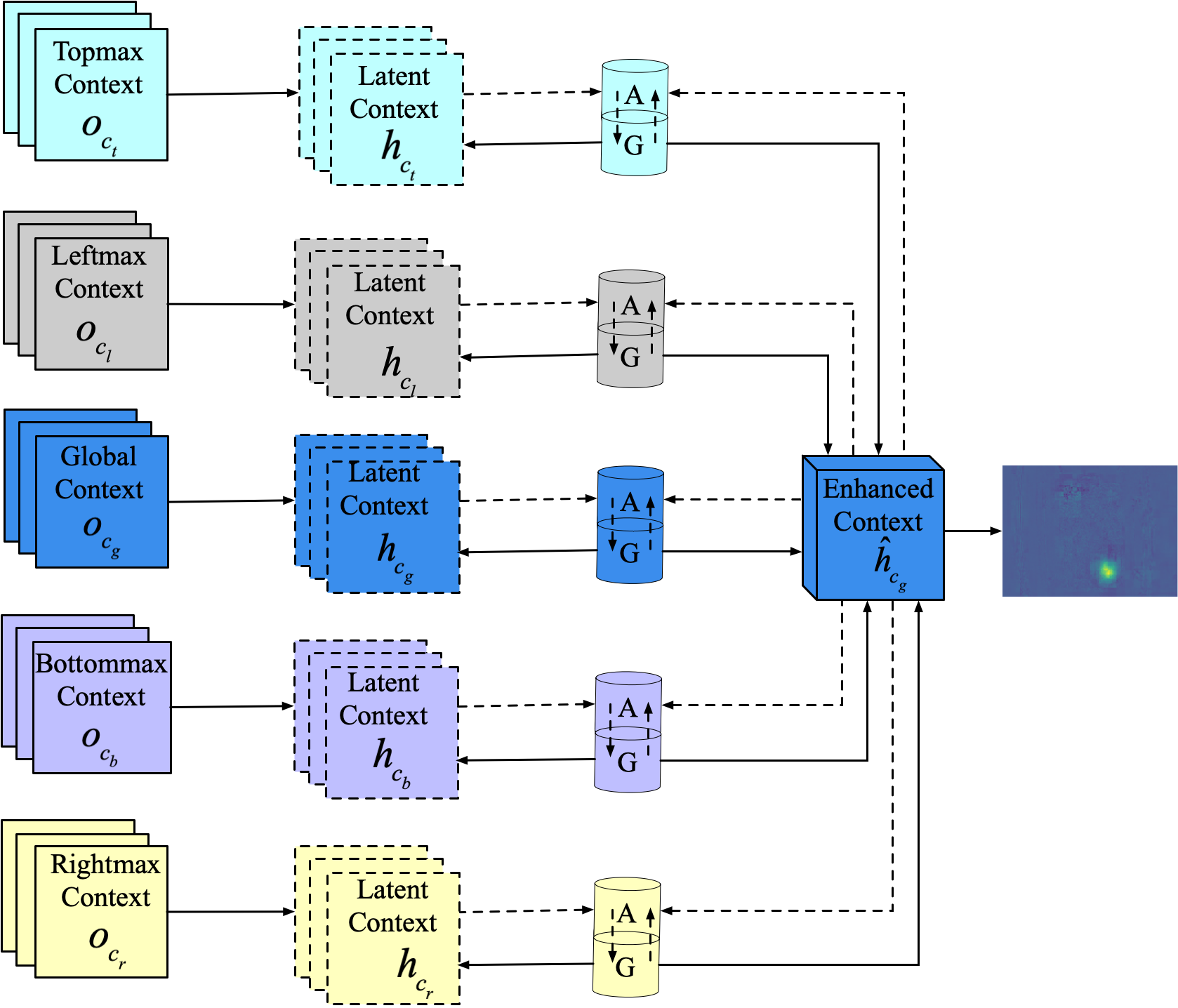}\\
(a) &(b)
\end{tabular}
\end{center}
\caption{Structured Context Mixer (SCM). (a) The whole structure of SCM. (b) The detailed architecture of our proposed graphical model. The arrows indicate the dependency between the variables used in our message passing algorithm, as shown in Algorithm \ref{Alg:mean-field}. The solid arrows denote the updates from the keypoint-specific and global contextual feature maps, while the dashed ones represent the updates from the two attention gates.}
\label{fig:SCM}
\end{figure*}

Although the global contextual information defined in Eq. (\ref{equation:global context}) can be used to directly infer the locations of keypoints, it pays less attention onto the local characteristics of each keypoint, i.e., keypoint-specific contextual information for describing the differences between body parts. Thus, in this subsection, we aim at modelling the dependency between global and keypoint-specific contextual information. Fig. \ref{fig:SCM} shows the structure of a graphical model based Structured Context Mixer. In Fig. \ref{fig:SCM}(a), ${O_{hg}}$ is the fusion of the output from multiple Hourglass modules, which preserve the multi-level feature representations. Here, we use five 3$ \times$3 Conv-BN-Relu layers to further refine the generated feature maps. One of them is considered as global contextual feature, whilst the other four maps are fed into Topmax, Leftmax, Bottommax and Rightmax shown in Fig. \ref{fig:maxdirection} respectively to generate keypoint-specific contextual features. Fig. \ref{fig:SCM}(b) shows the proposed graphical model which seamlessly fuses two types of contextual features for describing each keypoint. In particular, we design two attention gates ($A$ and $G$), which interact with each other to control the flow of the information between the coherent feature maps, i.e., global and keypoint-specific contextual feature maps.

 Given the collected contextual features $O$, we aim to infer the hidden structured contextual feature representation $H$ and the attention variables $A$ and $G$. We formalise the problem within a conditional random field framework and define a graphical model by constructing the following conditional distribution:
\begin{equation}
\begin{split}
P(H,A,G\left| I \right.,\Theta ) = \frac{1}{{Z(I,\Theta )}}\tilde P(H,A,G\left| I \right.,\Theta ){\rm{ }}
\label{equation:crf}
\end{split}
\end{equation}
where $I$ represents the input image, $Z(I,\Theta ) = Z$ is the partition function for normalisation, $\Theta$ is the set of parameters and $\tilde P =\exp \{  - E(H,A,G,I,\Theta)\} $. The energy function $E$ is defined as:
\begin{equation}
\begin{split}
E(H,A,G,I, \Theta) = \Phi (H,O) + \Psi (H,A) + \Gamma (H,G)
\label{equation:energy}
\end{split}
\end{equation}

The first term shown in Eq. (\ref{equation:energy}) denotes the unary potentials related to the variables $o_{c}^{i}$ and the corresponding latent variables $h_c^i$. It can be further defined as:
\begin{equation}
\begin{split}
\Phi (H,O) = \sum\limits_{c \in C} {\sum\limits_{i = 1}^N {\phi (h_c^i,o_c^i)} }
\label{equation:HO}
\end{split}
\end{equation}
According to the previous work \cite{xu2017learning}, we wish to represent the latent features. Thus, we adopt a multidimensional Gaussian distribution to represent $\phi (h_c^i,o_c^i)$ as follows:
\begin{equation}
\begin{split}
\phi (h_c^i,o_c^i) =  - \frac{1}{2}{(h_c^i - o_c^i)^T}{\Sigma ^{ - 1}}(h_c^i - o_c^i)
\label{equation:phi}
\end{split}
\end{equation}
where $\phi (h_c^i,o_c^i)$ is the Manhattan Distance between $h_c^i$ and $o_c^i$. $\sum {}$ is covariance matrix. Here, we assume $\sum {}$ is Identity Matrix $I$. The second and third terms shown in Eq. (\ref{equation:energy}) are two branches to model the relationship between the latent keypoint-specific and the latent global contextual feature maps. For each branch, we design a gate to regulate the flow of the information between the two types of the contextual features. Inspired by \cite{xu2017learning, chu2017multi}, we define two bilinear potentials, i.e. ${\psi _h}$ and ${\zeta _h}$, to represent the dependency between the latent keypoint-specific and the latent global contextual feature maps. $\Psi(H, A)$  and $\Gamma (H,G)$ can be defined as:
\begin{equation}\small
\begin{split}
\Psi (H,A) &= \sum\limits_{c \ne {c_g}} {\sum\limits_{i,j = 1}^N {\psi (a_c^i,h_c^i,h_{{c_g}}^j)} }= \sum\limits_{c \ne {c_g}} {\sum\limits_{i,j = 1}^N {a_c^i} } {\psi _h}(h_c^i,h_{{c_g}}^j)
\label{equation:HA}
\end{split}
\end{equation}
\begin{equation}\small
\begin{split}
\Gamma (H,G) &= \sum\limits_{c \ne {c_g}} {\sum\limits_{i,j = 1}^N {\zeta (g_c^i,h_c^i,h_{{c_g}}^j)} } =\sum\limits_{c \ne {c_g}} {\sum\limits_{i,j = 1}^N {g_c^i} } {\zeta _h}(h_c^i,h_{{c_g}}^j)
\label{equation:HG}
\end{split}
\end{equation}
where ${\psi _h}(h_c^i,h_{{c_g}}^j) = h_c^i\lambda _{i,j}^ch_{{c_g}}^j$ , ${\zeta _h}(h_c^i,h_{{c_g}}^j) = h_c^i\varpi _{i,j}^ch_{{c_g}}^j$, $
\lambda_{i, j}^{c}, \varpi_{i, j}^{c} \in \mathbb{R}^{L_{c} \times L_{c_{g}}}$, and ${L_c}$ and ${L_{{c_g}}}$ denote the number of the channels of the keypoint-specific and global contextual feature maps, respectively.

Using two branches with different gates, shown in Eqs. (\ref{equation:HA}) and (\ref{equation:HG}), we also consider the relationship between the two gates. Different from previous works that model the spatial relationship between pixels within a feature map \cite{chu2017multi, xu2018structured }, we highlight the spatial dependency between feature maps. In our algorithm, the two gates are spatially interactive, guiding our proposed GM-SCENet to preserve useful keypoint-specific contextual information. We define:
\begin{equation}\small
\begin{split}
\Xi (A,G) &= \sum\limits_{c \ne {c_g}} {\sum\limits_{i,j = 1}^N {\vartheta (a_c^i,g_c^j)} } =\sum\limits_{c \ne {c_g}} {\sum\limits_{i,j = 1}^N {a_c^i\eta _{i.j}^cg_c^j} }
\label{equation:AG}
\end{split}
\end{equation}
where $\eta_{i . j}^{c} \in \mathbb{R}^{L_{c}^{a} \times L_{c}^{g}}$, $L_c^a$, $L_c^g$ are the numbers of the channels of gate $A$ and $G$, respectively.

To deduce $P(H,A,G\left| I \right.,\Theta )$ shown in Eq. (\ref{equation:crf}), we adopt classical mean-field approximation \cite{krahenbuhl2011efficient} which is effective and efficient to handle high-dimensional data. $P(H,A,G\left| I \right.,\Theta )$ can be approximated by a new distribution $Q(H,A,G\left| {I,\Theta } \right.) $, which is represented as a product of independent distributions as follows:
\begin{equation}
\begin{split}
Q(H,A,G\left| I \right.,\Theta ) = \prod\limits_{i = 1}^N {{q_i}(h_c^i){q_i}(h_{{c_g}}^i){q_i}(a_c^i){q_i}(g_c^i)}
\label{equation:P}
\end{split}
\end{equation}
where ${q_i}(h_c^i)$, ${q_i}(h_{{c_g}}^i)$, ${q_i}(a_c^i)$ and ${q_i}(g_c^i)$ are independent marginal distributions. Afterwards, we minimise the differences between the two distributions using Kullback–Leibler (KL) divergence, which is formulated as follows:
\begin{equation}\small
\begin{split}
\begin{array}{l}
{D_{KL}}[Q(H,A,G\left| {I,\Theta } \right.)||P(H,A,G\left| {I,\Theta } \right.)]\\
 = \sum\limits_x {Q(x)} \log Q(x) - \sum\limits_x {Q(x)} \log P(x)\\
 = \sum\limits_x {Q(x)} \log Q(x) - \sum\limits_x {Q(x)} \log \tilde P + \sum\limits_x {Q(x)} \log Z
\end{array}
\label{equation:DKL}
\end{split}
\end{equation}

Our target is to minimise the right hand side of the above equation. We convert Eq. (\ref{equation:DKL}) to the following form:
\begin{equation}\small
\begin{split}
\begin{array}{c}
L\left(h_{c}^{i}\right)=\sum\limits_{h_{c}^{i}} q_{i}\left(h_{c}^{i}\right) \log \left(q_{i}\left(h_{c}^{i}\right)\right)- \\ \sum\limits_{h_{c}^{i}} q_{i}\left(h_{c}^{i}\right) \mathbb{E}_{\bar{q}_{i}\left(h_{c}^{i}\right)}\{\log \tilde{P}\}+C
\end{array}
\label{equation:Lhci}
\end{split}
\end{equation}
\begin{equation}\small
\begin{split}
{\bar q_i}(h_c^i) = {q_i}\left( {h_{{c_g}}^i} \right){q_i}\left( {a_c^i} \right){q_i}\left( {g_c^i} \right) \cdot \\\prod\limits_{j \ne i} {{q_j}\left( {h_c^j} \right){q_j}\left( {h_{{c_g}}^j} \right){q_j}\left( {a_c^j} \right){q_j}\left( {g_c^j} \right)}
\label{equation:qba}
\end{split}
\end{equation}
where $\mathbb{E}_{\bar{q}_{i}\left(h_{c}^{i}\right)}$ represents the expectation of distribution ${\bar q_i}(h_c^i)$, and $C$ is the constant. Then, we minimise $L(h_c^i)$, which can be regarded as a constrained optimisation problem. Given the condition $\int {{q_i}(h_c^i)d} h_c^i = 1$, we rewrite Eq. (\ref{equation:Lhci}) as:
\begin{equation}\small
\begin{split}
\begin{aligned}
&\arg \min _{q_{i}}\left(L\left(h_{c}^{i}\right)\right)= \arg \min _{q_{i}}\left(\int q_{i}\left(h_{c}^{i}\right) \log q_{i}\left(h_{c}^{i}\right) d h_{c}^{i}-\right.\\ &\left.\int q_{i}\left(h_{c}^{i}\right) \mathbb{E}_{\bar{q}_{i}\left(h_{c}^{t}\right)}\{\log \tilde{P}\} d h_{c}^{i}+\lambda_{i}\left(\int q_{i}\left(h_{c}^{i}\right) d h_{c}^{i}-1\right)\right)
\end{aligned}
\label{equation:argmin}
\end{split}
\end{equation}

To seek the minimum, we take the derivative of $L(h_c^i)$  with respect to ${q_i}(h_c^i)$  and set the derivative to zero, as shown in Eq. (\ref{equation:derivative}). To achieve the minimisation, we can derive the  mean-field update shown in Eq. (\ref{equation:hci}) by combining Eqs. (\ref{equation:crf})-(\ref{equation:P}).
\begin{equation}\small
\begin{split}
\begin{aligned}
\frac{\partial L\left(h_{c}^{i}\right)}{\partial q_{i}\left(h_{c}^{i}\right)}=\log q_{i}\left(h_{c}^{i}\right)-\mathbb{E}_{\bar{q}_{i}\left(h_{c}^{i}\right)}\{\log \tilde{P}\}+\lambda_{i}=0
\end{aligned}
\label{equation:derivative}
\end{split}
\end{equation}
\begin{equation}\small
\begin{split}
\begin{aligned}
q_{i}\left(h_{c}^{i}\right) &\propto \exp \bigg(\phi\left(h_{c}^{i}, o_{c}^{i}\right)+\\
&\mathbb{E}_{q_{i}\left(a_{c}^{i}\right)}\left\{a_{c}^{i}\right\} \sum_{j=1}^{N} \mathbb{E}_{q_{i}\left({h}_{c_{g}}^{j}\right)}\left\{\psi_{h}\left(h_{c}^{i}, h_{c_{g}}^{j}\right)\right\}+\\ &\mathbb{E}_{q_{i}\left(g_{c}^{i}\right)}\left\{g_{c}^{i}\right\} \sum_{j=1}^{N} \mathbb{E}_{q_{i}\left({h}_{c_{g}}^{j}\right)}\left\{\zeta_{h}\left(h_{c}^{i}, h_{c_{g}}^{j}\right)\right\}\bigg)
\end{aligned}
\label{equation:hci}
\end{split}
\end{equation}

Eqs. (\ref{equation:Lhci})-(\ref{equation:hci}) show the process of mean-field approximation where latent variables are updated, while we fix the remaining latent variables. Following this way, ${q_i}(h_{{c_g}}^i)$ , ${q_i}(a_c^i)$, ${q_i}(g_c^i)$ can be derived as:
\begin{equation}\small
\begin{split}
\begin{aligned}
 q_{i}\left(h_{c_{g}}^{i}\right) &\propto \exp \bigg(\phi\left(h_{c_{g}}^{i}, o_{c_{g}}^{i}\right)+\\
 &\sum_{c \neq c_{g}} \sum_{j=1}^{N} \mathbb{E}_{q_{i}\left(a_{c}^{j}\right)}\left\{a_{c}^{j}\right\} \mathbb{E}_{q_{i}\left(h_{c}^{j}\right)}\left\{\psi_{h}\left(h_{c_{g}}^{i}, h_{c}^{j}\right)\right\}+\\
 &\sum_{c \neq c_{g}} \sum_{j=1}^{N} \mathbb{E}_{q_{i}\left(g_{c}^{j}\right)}\left\{g_{c}^{j}\right\} \mathbb{E}_{q_{i}\left(h_{c}^{j}\right)}\left\{\zeta_{h}\left(h_{c_{g}}^{i}, h_{c}^{j}\right)\right\}\bigg)
\end{aligned}
\label{equation:hcgi}
\end{split}
\end{equation}
\begin{equation}\small
\begin{split}
\begin{aligned}
q_{i}\left(a_{c}^{i}\right) &\propto \exp \bigg(a_{c}^{i} \mathbb{E}_{q_{i}\left(h_{c}^{i}\right)}\left\{\sum_{j=1}^{N} \mathbb{E}_{q_{i}\left(h_{\varepsilon_{g}}^{j}\right)}\left\{\psi_{h}\left(h_{c}^{i}, h_{c_{g}}^{j}\right)\right\}\right\}+\\
&\sum_{c \neq c_{g}} \sum_{j=1}^{N} \mathbb{E}_{q_{i}\left(g_{c}^{j}\right)}\left\{\vartheta\left(a_{c}^{i}, g_{c}^{j}\right)\right\}\bigg)
\end{aligned}
\label{equation:aci}
\end{split}
\end{equation}

\begin{equation}\small
\begin{split}
\begin{aligned}
q_{i}\left(g_{c}^{i}\right) &\propto \exp \bigg(g_{c}^{i} \mathbb{E}_{q_{i}\left(h_{c}^{i}\right)}\left\{\sum_{j=1}^{N} \mathbb{E}_{q_{i}\left(h_{c_{g}}^{j}\right)}\left\{\zeta_{h}\left(h_{c}^{i}, h_{c_{g}}^{j}\right)\right\}\right\}+\\
&\sum_{c \neq c_{g}} \sum_{j=1}^{N} \mathbb{E}_{q_{i}\left(a_{c}^{j}\right)}\left\{\vartheta\left(g_{c}^{i}, a_{c}^{j}\right)\right\}\bigg)
\end{aligned}
\label{equation:gci}
\end{split}
\end{equation}

Eqs. (\ref{equation:hci}) and (\ref{equation:hcgi}) show the posterior distributions of $h_c^i$ and $h_{{c_g}}^i$ in Gaussian distributions. We denote the expectation of $h_c^i$, $h_{{c_g}}^i$, $a_c^i$ and $g_c^i$ by $\hat{h}_{c}^{i}=\mathbb{E}_{q_{i}\left(h_{c}^{i}\right)}\left\{h_{c}^{i}\right\}$, $\hat{h}_{c_{g}}^{i}=\mathbb{E}_{q_{i}\left(h_{c_{g}}^{i}\right)}\left\{h_{c_{g}}^{i}\right\}$, $\hat{a}_{c}^{i}=\mathbb{E}_{q_{i}\left(a_{c}^{i}\right)}\left\{a_{c}^{i}\right\}$ and $\hat{g}_{c}^{i}=\mathbb{E}_{q_{i}\left(g_{c}^{i}\right)}\left\{g_{c}^{i}\right\}$  respectively. By combining Eqs. (\ref{equation:phi}), (\ref{equation:HA}) and (\ref{equation:HG}), we derive the following mean-field updates for the latent structured contextual representations:
\begin{equation}
\begin{split}
\begin{aligned}
\hat h_c^i = o_c^i + \hat a_c^i\sum\limits_{j = 1}^N {\lambda _{i,j}^c} \hat h_{{c_g}}^j + \hat g_c^i\sum\limits_{j = 1}^N {\varpi _{i,j}^c} \hat h_{{c_g}}^j
\end{aligned}
\label{equation:Exphci}
\end{split}
\end{equation}
\begin{equation}
\begin{split}
\begin{aligned}
\hat h_{{c_g}}^i = o_{{c_g}}^i + \sum\limits_{c \ne {c_g}} {\sum\limits_{j = 1}^N {\hat a_c^j\lambda _{i,j}^c} \hat h_c^j + \sum\limits_{c \ne {c_g}} {\sum\limits_{j = 1}^N {\hat g_c^j\varpi _{i,j}^c} \hat h_c^j} }
\end{aligned}
\label{equation:Exphcgi}
\end{split}
\end{equation}

\begin{algorithm}[H]
\caption{Algorithm for mean-field updates in our proposed Structured Context Mixer with CNN.}
\begin{algorithmic}[1]
\renewcommand{\algorithmicrequire}{\textbf{Input:}}
\renewcommand{\algorithmicensure}{\textbf{Output:}}
\REQUIRE Global contextual feature maps ${{\hat h}_{{c_g}}}$ initialised with corresponding feature observation ${o_{{c_g}}}$, keypoint-specific contextual feature maps ${{\hat h}_{{c}}}$ initialized with  ${o_{{c}}}$, and the number of iteration $N$.
\ENSURE Enhanced structured contextual feature maps, i.e., global contextual feature maps generated in the last iteration ${{\hat h}_{{c_g}}}$ shown in Fig. \ref{fig:SCM}(b) and Eq. (\ref{equation:Exphcgi}).
\FOR {$i \leftarrow 1 $ to $ N $}
 \STATE ${{\tilde g}_c} \leftarrow {{\hat h}_c} \odot ({\varpi _c} \otimes {{\hat h}_{{c_g}}})$, ${{\tilde a}_c} \leftarrow {{\hat h}_c} \odot ({\lambda _c} \otimes {{\hat h}_{{c_g}}})$;
 \STATE ${{\mathord{\buildrel{\lower3pt\hbox{$\scriptscriptstyle\frown$}}
\over g} }_c} \leftarrow {\eta _c} \otimes {{\tilde a}_c}$, ${{\tilde a}_c} \leftarrow {\eta _c} \otimes {{\tilde g}_c}$;
 \STATE ${{\hat g}_c} \leftarrow \sigma ( - ({{\tilde g}_c} \oplus {{\mathord{\buildrel{\lower3pt\hbox{$\scriptscriptstyle\frown$}}
\over g} }_c}))$;
 \STATE ${{\hat a}_c} \leftarrow \sigma ( - ({{\tilde a}_c} \oplus {{\mathord{\buildrel{\lower3pt\hbox{$\scriptscriptstyle\frown$}}
\over a} }_c}))$;
 \STATE ${{\tilde h}_{{c_g}}} \leftarrow \sum\limits_{c \ne {c_g}} {{{\hat g}_c} \odot (} {\varpi _c} \otimes {{\hat h}_c})$;
 \STATE ${{\mathord{\buildrel{\lower3pt\hbox{$\scriptscriptstyle\frown$}}
\over h} }_{{c_g}}} \leftarrow \sum\limits_{c \ne {c_g}} {{{\hat a}_c} \odot (} {\lambda _c} \otimes {{\hat h}_c})$;
 \STATE ${{\hat h}_{{c_g}}} \leftarrow {o_{{c_g}}} \oplus {{\tilde h}_{{c_g}}} \oplus {{\mathord{\buildrel{\lower3pt\hbox{$\scriptscriptstyle\frown$}}
\over h} }_{{c_g}}}$;
 \STATE ${{\tilde h}_c} \leftarrow {{\hat g}_c} \odot ({\varpi _c} \otimes {{\hat h}_{{c_g}}})$;
 \STATE ${{\mathord{\buildrel{\lower3pt\hbox{$\scriptscriptstyle\frown$}}
\over h} }_c} \leftarrow {{\hat a}_c} \odot ({\lambda _c} \otimes {{\hat h}_{{c_g}}})$;
 \STATE ${{\hat h}_c} \leftarrow {o_c} \oplus {{\tilde h}_c} \oplus {{\mathord{\buildrel{\lower3pt\hbox{$\scriptscriptstyle\frown$}}
\over h} }_c}$;
\ENDFOR
\RETURN Enhanced structured contextual feature maps ${{\hat h}_{{c_g}}}$.
\end{algorithmic}
\label{Alg:mean-field}
\end{algorithm}

In Section \ref{section3.1.2}, we define $a_c^i$ and $g_c^i$ as binary variables. Therefore, the expectations of their distributions are the same as $q(a_c^i = 1)$ and $q(g_c^i = 1)$ respectively. The expectations $\hat a_c^i$ and $\hat g_c^i$ can be derived using the distributions shown in Eqs. (\ref{equation:aci}) and (\ref{equation:gci}) and the potential functions defined in Eqs. (\ref{equation:HA}), (\ref{equation:HG}) and (\ref{equation:AG}):
\begin{equation}
\begin{split}
\begin{aligned}
\hat a_c^i &= \sigma ({\rm{ - }}\sum\limits_{j = 1}^N {\hat h_c^i\lambda _{i,j}^c} \hat h_{{c_g}}^j - \sum\limits_{c \ne {c_g}} {\sum\limits_{j = 1}^N {\eta _{i,j}^c} \hat g_c^j} )
\end{aligned}
\label{equation:Expaci}
\end{split}
\end{equation}
\begin{equation}
\begin{split}
\begin{aligned}
\hat g_c^i &= \sigma ({\rm{ - }}\sum\limits_{j = 1}^N {\hat h_c^i\varpi _{i,j}^c} \hat h_{{c_g}}^j - \sum\limits_{c \ne {c_g}} {\sum\limits_{j = 1}^N {\eta _{i,j}^c} \hat a_c^j} )
\end{aligned}
\label{equation:Expgci}
\end{split}
\end{equation}
where $\sigma(\cdot)$ is the sigmoid function. Using Eq. (\ref{equation:Expgci}), the updates for gate $g_c^i$ depend on the values of the hidden features $h_c^i$ and the other gate $a_c^i$. In particular, the spatially interactive gates allow us to adjust the distribution of the hidden features, i.e., keypoint-specific and the global contextual features shown in Eqs. (\ref{equation:Exphci}) and (\ref{equation:Exphcgi}), respectively.

\subsubsection{End-to-End Optimisation}
\label{section3.1.3}
Following \cite{xu2017learning}, \cite{chu2016crf}, we convert the mean-field inference equations shown in Eqs. (\ref{equation:Exphci})-(\ref{equation:Expgci}) to convolutional operations by deep neural networks for jointly training our Structured Context Mixer and the network. Our goal is to achieve end-to-end optimisation of the proposed GM-SCENet.

The mean-field updates of the two gates shown in Eqs. (\ref{equation:Expaci}) and (\ref{equation:Expgci}) can be implemented with deep neural networks in several steps as follows: (a) Message passing between the global contextual feature maps ${\hat h_{{c_g}}}$ and the keypoint-specific contextual feature maps ${\hat h_c}$ can be performed by ${{\tilde g}_c} \leftarrow {{\hat h}_c} \odot ({\varpi _c} \otimes {{\hat h}_{{c_g}}})$ and ${{\tilde a}_c} \leftarrow {{\hat h}_c} \odot ({\lambda _c} \otimes {{\hat h}_{{c_g}}})$, where ${\varpi _c}$ and ${\lambda _c}$ represents the convolutional kernels and the symbols $\odot$ and $ \otimes$ denote the element-wise product and convolutional operation, respectively; (b) Message passing between the two gates is performed by ${{\mathord{\buildrel{\lower3pt\hbox{$\scriptscriptstyle\frown$}}
\over g} }_c} \leftarrow {\eta _c} \otimes {{\tilde a}_c}$ and ${{\tilde a}_c} \leftarrow {\eta _c} \otimes {{\tilde g}_c}$, where ${\eta _c}$ is a convolution kernel; (c) The normalisation of the two gates is performed by ${{\hat g}_c} \leftarrow \sigma ( - ({{\tilde g}_c} \oplus {{\mathord{\buildrel{\lower3pt\hbox{$\scriptscriptstyle\frown$}}
\over g} }_c}))$ and ${{\hat a}_c} \leftarrow \sigma ( - ({{\tilde a}_c} \oplus {{\mathord{\buildrel{\lower3pt\hbox{$\scriptscriptstyle\frown$}}
\over a} }_c}))$, where $ \oplus$ represents the element-wise addition operation.

Afterwards, we conduct mean-field updating of the global contextual feature maps ${h_{{c_g}}}$ shown in Eq. (\ref{equation:Exphcgi}) after having updated the two gates. This process goes through: (a) Message passing from the keypoint-specific to the global contextual feature maps under the control of gate ${g_c}$ is achieved by ${{\tilde h}_{{c_g}}} \leftarrow \sum\limits_{c \ne {c_g}} {{{\hat g}_c} \odot (} {\varpi _c} \otimes {{\hat h}_c})$; (b) Message passing from the keypoint-specific to the global contextual feature maps under the control of gate ${a_c}$ is performed by ${{\mathord{\buildrel{\lower3pt\hbox{$\scriptscriptstyle\frown$}}
\over h} }_{{c_g}}} \leftarrow \sum\limits_{c \ne {c_g}} {{{\hat a}_c} \odot (} {\lambda _c} \otimes {{\hat h}_c})$. (c) The final update involves unary term, i.e., the observed global contextual feature maps ${o_{{c_g}}}$ can be performed by ${{\hat h}_{{c_g}}} \leftarrow {o_{{c_g}}} \oplus {{\tilde h}_{{c_g}}} \oplus {{\mathord{\buildrel{\lower3pt\hbox{$\scriptscriptstyle\frown$}}
\over h} }_{{c_g}}}$.

 Similarly, following Eq. (\ref{equation:Exphci}), the updating of the latent feature maps corresponding to the keypoint-specific contextual information is straightforward. The mean-field updates in our proposed SCM are summarised in Algorithm \ref{Alg:mean-field}.

\subsection{Cascaded Multi-Level Supervision}
\label{section3.2}
\begin{figure}
\begin{center}
\includegraphics[width=7.8cm]{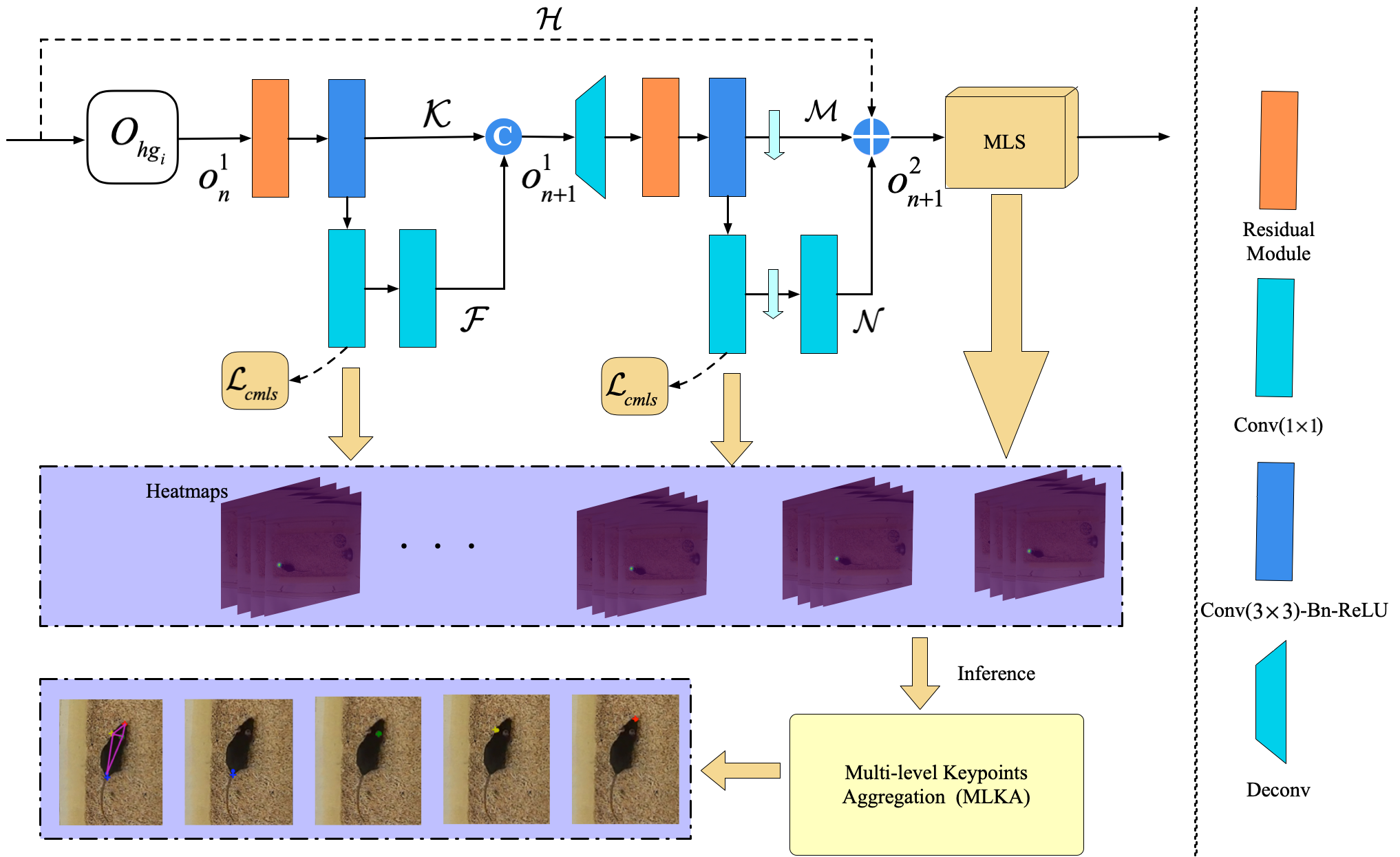}
\end{center}
\caption{Architecture of the Cascaded Multi-level Supervision module (CMLS).}
\label{fig:CMLS}
\end{figure}

Since mice are small and highly deformable, single-scale supervision cannot work stably in different set-ups \cite{newell2016stacked}, \cite{Mathis2018}. In this section, we design a novel Cascaded Multi-level Supervision module (CMLS) consisting of two identical Multi-level Supervision blocks (MLS) as intermediate supervision to jointly train the proposed SCM and the Hourglass network, where multi-level features, i.e., multi-stage features of multiple scales, are used for supervised learning, as shown in Figs. \ref{fig:framework} and \ref{fig:CMLS}. HRNet\cite{Sun2019} also includes large and small scale representations, but this model performs multi-scale fusion among sub-networks without considering the multi-scale features for supervised learning. On the contrary, our MLS block provides a coarse-to-fine supervision to strengthen multi-scale feature learning. Additionally, different from \cite{ke2018multi}, our CMLS module is used for module-wise supervision rather than layer-wise supervision. In other words, we adopt this module to connect different Hourglass modules and further refine the prediction results from SCM, as shown in Fig. \ref{fig:framework}. Particularly, for each MLS block, we combine the supervision information with historical information by the $Concat$ operation in the first stage, before adopting a $Deconv$ layer (Deconvolution) to extract high-resolution information in the second stage. We use the $Concat$ here to preserve feature maps of higher dimensions. This allows the $Deconv$ operation to capture more high-resolution information, which can provide more detailed features, thus benefiting the large-scale supervision in the second stage. Similar to \cite{newell2016stacked}, we also use identity mapping to add the input information to the output of the second stage. Furthermore, two MLS blocks are concatenated to bulid CMLS module where we take the output of the first MLS block $o_{n + 1}^2$ as the input of the other MLS block. In this way, small-scale supervision information is used as prior knowledge for large-scale supervision, and then the large-scale supervision information is also considered as prior knowledge of the following MLS module. In addition, the generated multi-level information can drive our SCM to strengthen the contextual feature learning. Therefore, such coarse-to-fine process can help our GM-SCENet to generate multi-level feature maps for precisely localising the keypoints of the mouse.

The mathematical formulation of the first and second stages in our proposed CMLS are as follows:
\begin{equation}
\begin{split}
o_{n + 1}^1 = Concat({\cal K}(o_n^1,W_n^{\cal K}),{\cal F}(o_n^1,W_n^{\cal F}))
\label{equation: stage1}
\end{split}
\end{equation}
\begin{equation}
\begin{split}
o_{n + 1}^2 = {\cal H}\left( {o_n^2} \right) + {\cal M}(o_n^2,W_n^{\cal M}) + {\cal N}(o_n^2,W_n^{\cal N})
\label{equation:stage2}
\end{split}
\end{equation}
where $concat( \cdot )$ denotes the channel-wise concatenation, $o_n^1$ and $o_{n + 1}^1$ are the input and output of the first stage of the $n$-th MLS module. $o_n^2$ and $o_{n + 1}^2$ are the input and output of the second stage of the $n$-th MLS module and $o_{n + 1}^1 = o_n^2$. ${\cal H}\left( {o_n^2} \right) = o_n^2$ refers to the identity mapping. ${\cal K}(o_n^1,W_n^{\cal K})$ is a stack of the Residual Module and Conv($3 \times 3$)-Bn-ReLU. ${\cal F}(o_n^1,W_n^{\cal F})$ represents two $1 \times 1$ convolutions. ${\cal M}(o_n^2,W_n^{\cal M})$ is a stack of  deconvolution and residual blocks followed by Conv($3 \times 3$)-Bn-ReLU and downsampling operation. ${\cal N}(o_n^2,W_n^{\cal N})$ is a combination of $1 \times 1$ convolution, downsampling operation and $1 \times 1$ convolution.

We use the Mean-Squared Error (MSE) based loss function \cite{Graving2019}, \cite{chu2017multi} for the training of the entire network. To represent the ground-truth keypoint, we generate a heatmap $h_k$ for each single keypoint $k$, $k \in \{ 1,2,...,K\}$ by a 2D Gaussian distribution centered at the mouse part position ${l_k} = ({x_k},{y_k})$.
We apply a MSE loss function to the proposed SCM and the CMLS module (i.e., ${{\cal L}_{scm}}$ and ${{\cal L}_{cmls}}$ in Fig. \ref{fig:framework}) as follows:
\begin{equation}
\begin{split}
\mathcal{L}=\sum_{k=1}^{K} \sum_{p \in \mathbb{P}}|| \hat{h}_{k}(p)-h_{k}(p)||_{2}^{2}
\label{equation:mse}
\end{split}
\end{equation}
where $p \in \mathbb{P}=\{(x, y) \mid x=1, \ldots, W, y=1, \ldots, H\}$  denotes the $p$-th location, ${\hat h_k}$ denotes the predicted heatmap of keypoint $k$. In our system, we add all the supervision losses from SCM and the CMLS module. During the inference phase, we obtain the location of the keypoint  from the predicted heatmap ${\hat h_k}$ by choosing the position with the maximum score, i.e., ${\hat l_k} = \arg \max {\hat h_k}$. Then, combining all the predicted heatmaps from the inner layers of the GM-SCENet and using Multi-Level Keypoint Aggregation introduced in Section \ref{section3.3}, we have the final prediction results.

\subsection{Multi-level Keypoints Aggregation for Inference}
\label{section3.3}
In general, the final prediction results of deep models for pose estimation come from the last stage of the networks. Here, we argue that results generated in the inner layers may be more accurate if we apply intermediate supervision including our proposed CMLS module. Similar to the popular Non-Maximum Suppression (NMS) postprocessing step used in object detection, most top-down multi-person pose estimation also adopts this method to remove unreasonable poses \cite{xiao2018simple}. Inspired by this concept, we propose an inference algorithm for single-mouse pose estimation, i.e., Multi-level Keypoint Aggregation (MLKA), to integrate  prediction results generated by SCM and the CMLS module. We adopt the CMLS module as intermediate supervision for training our proposed framework, and consider all the predicted heatmaps with multiple scales as potential candidates, as shown in Fig. \ref{fig:CMLS}.

During the inference phase, we first obtain the positions of the keypoints from the generated heatmaps before adopting MLKA to calculate the average heatmaps of the selected multi-level keypoints, i.e., multi-scale keypoints from different CMLS modules. Specifically, we change the coordinates of different keypoints in proportion to the corresponding image sizes before choosing the keypoint candidates which are close to those from the last stage. We then use the Object Keypoint Similarity (OKS) metric to examine the similarities of the keypoints from the same part. After having selected all the  potential keypoints of different parts by the preset threshold, we aggregate all the selected keypoints of body parts by directly calculating the average of them to generate a new set of keypoints to reconstruct the skeleton. The OKS metric has been used for human pose estimation in different datasets \cite{Sun2019}. Therefore, we use the modified OKS metric to measure the similarity between the results generated in the last stage and those predicted values in the inner layers. The modified OKS can be formed as:
\begin{equation}
\begin{split}
OKS = \frac{{\sum\limits_i {\exp } \left( { - d_i^2/k_i^2} \right)\delta \left( {{v_i} > 0} \right)}}{{\sum\limits_i \delta  \left( {{v_i} > 0} \right)}}
\label{equation:oks}
\end{split}
\end{equation}
where $d_i^2$ is the Euclidean distance between the detected $i$-th keypoint and the corresponding ground truth, and $\delta ( \cdot )$ is the Kronecker function. $\delta ( \cdot ) = 1$ if the condition holds, otherwise 0. ${v_i}$ is the visibility flag of the ground truth, and ${k_i}$ is a per-keypoint constant that controls the falloff.

\section{Experimental Setup}
To evaluate the performance of our proposed methods, we implement comprehensive evaluations on two mouse pose datasets (i.e., DeepLabCut Mouse Pose\footnote{https://zenodo.org/record/4008504${\rm{\# }}$.X1KM7mczYW8} and our PDMB datasets) and other two animal datasets (i.e., Zebra \cite{Graving2019} and Fly \cite{Pereira2019}) and a human pose dataset (i.e., LSP \cite{johnson2010clustered}). In this section, we first introduce the two mouse datasets and evaluation metrics. Then, we describe the implementation details.

\subsection{Datasets and Evaluation Metrics}
\label{section4.1}
\subsubsection{DeepLabCut Mouse Pose Dataset}
\label{section4.1.1}
DeepLabCut Mouse Pose dataset: The dataset is  recorded by  two  different cameras with the resolution of 640*480 and 1700*1200 pixels respectively. Most images are 640*480, while some images were cropped around mice to generate the images that are approximately 800*800. There are 1066 images from multiple experimental sessions of 7 different mice. We randomly split the dataset into a training set of 853 images and a test set of 213 images. In addition, four parts (i.e., snout, left ear, right ear and tail base) are labeled in this dataset.

\subsubsection{PDMB Dataset}
\label{section4.1.2}
In this paper, we introduce a new dataset that is collected in collaboration with the biologists of Queen’s University Belfast of United Kingdom, for a study on motion recordings of mice with Parkinson’s disease (PD) \cite{jiang2021muti}. The neurotoxin 1-methyl-4-phenyl-1,2,3,6-tetrahydropyridine (MPTP) is used as a model of PD, which has become an invaluable aid to produce experimental parkinsonism since its discovery in 1983 \cite{jackson2007protocol}. All the mice receive treatment of MPTP and they are housed in a controlled environment with the constant temperature of (27$^{\circ}C$) and light condition (long fluorescent lamp of 40W), and under constant climatic conditions with free access to food and water (placed on the corner of the cage). All experimental procedures are performed in accordance with the Guidance on the Operation of the Animals (Scientific Procedures) Act, 1986 (UK) and approved by the Queen’s University Belfast Animal Welfare and Ethical Review Body.

We record 4 videos for 4 mice using a Sony Action camera (HDR-AS15) (top-view) with a frame rate of 30 fps and 640*480 resolution. For each video, We divide it into 6 clips, and each clip lasts about 10 minutes. Then, we extract frames at different frequencies and build our top-view PDMB dataset with 4248, 3000, 2000 images for training, validation and testing, respectively. Particularly, our dataset is more challenging than the existing DeepLabCut Mouse Pose dataset. For example, our PDMB dataset contains a wide range of behaviours \cite{jiang2018context} (e.g., rear, groom and eat), and abnormal poses. Fig.\ref{fig:PDMB} shows video frames for different behaviours. In addition, the PDMB dataset also provides an additional annotation indicates the lack of visibility. More details about the data annotation are described in Supplementary G.


\subsubsection{Evaluation Metrics}
\label{section4.1.3}
 Following \cite{Mathis2018}, we use Root Mean Square Error (RMSE) as an evaluation metric. Furthermore, we introduce a new evaluation metric, i.e., Percentage of Correct Keypoints (PCK) to the PDMB and DeepLabCut Mouse Pose datasets. Different from the PCK metric used for the datasets of human pose estimation, we modify PCK and apply it to our PDMB dataset, as shown in Eq. (\ref{equation:PCK}). A detected keypoint is considered correct if the normalized distance between the predicted and the ground-truth keypoints falls within a certain threshold $T$. In our experiments, $T$ is set to 0.2, and we also draw PCK curves with different thresholds to compare different approaches.
\begin{equation}
\begin{split}
PCK@T = \frac{{\sum\limits_k {\delta \left( {\left\| {\frac{{{{\hat l}_k} - {l_k}}}{{\vec c}}} \right\|_2^{} < T} \right)} }}{{\sum\limits_k 1 }}
\label{equation:PCK}
\end{split}
\end{equation}
where $T$ represents the threshold evaluating the normalized distance between ${\hat l_k}$ and ${l_k}$, $\vec c$ represents a constant vector for normalisation.

\subsection{Implementation Details}
\label{section4.2}
All the experiments are performed on a server with an Intel Xeon CPU @ 2.40GHz and two 16GB Nvidia Tesla P100 GPUs. The parameters are optimised by the Adam algorithm. For the DeepLabCut Mouse Pose dataset, we use the initial learning rate of 1e-4 and all the training and test images are resized to the resolution of 640*480. For the PDMB dataset, the initial learning rate is set to le-5. We adopt the validation split of the PDMB dataset to monitor the training process. The proposed GM-SCENet is implemented with Pytorch. The source code and dataset will be available at: \href{https://github.com/FeixiangZhou/GM-SCENet}{https://github.com/FeixiangZhou/GM-SCENet.}

\section{Results and Discussion}
In this section, we implement comprehensive experiments to evaluate the performance of our proposed methods. Firstly, to validate the effectiveness of the proposed SCM, CMLS module and MLKA, we conduct ablation experiments on the proposed PDMB validation and DeepLabCut Mouse pose test datasets. We use the 3-stack Hourglass network as our baseline network. Based on this, we first investigate each proposed component, followed by the comprehensive analysis for the impact of each module. Then, we compare our network with the other state-of-the-art networks on both PDMB and DeepLabCut Mouse pose test datasets.

\begin{table}[htbp]
\begin{center}
\caption{Ablation experiments for the Structured Context Mixer on the PDMB validation dataset (RMSE(pixels) and PCK@0.2).}
\label{tab:scm}
\begin{tabular}{p{2.6cm}p{0.5cm}p{0.5cm}p{0.5cm}p{0.5cm}p{0.5cm}p{0.5cm}}
\toprule
Methods\quad & RMSE & Snout &  Left ear & Right ear & Tail base & Mean\\
\midrule

Baseline  & 4.89 & 93.38  & 92.67 & 93.97 & 86.87 & 91.72\\
\hline

\multicolumn{7}{c}{\textit{SCM(Conv($3 \times 3$))}}\\
\hline
SCM(Iteration 1)  & 4.98 & 94.99  & 93.87 & 95.77 & 83.4 & 92.0\\
SCM(Iteration 2)  & 3.74 & 96.90  & 96.50 & 96.07 & 88.87 & 94.58\\
SCM(Iteration 3)  & \textbf{3.07} & 97.67 & \textbf{97.83}  & 97.17 & \textbf{91.67} & \textbf{96.08}\\
\hline
\multicolumn{7}{c}{\textit{SCM(Conv($1 \times 1$))}}\\
\hline
SCM(Iteration 1) & 4.79 & 92.55  & 86.83 & 93.60 & 85.83 & 89.70\\
SCM(Iteration 2)  & 3.91 & 95.13  & 91.97 & 96.90 & 88.93 & 93.23\\
SCM(Iteration 3)  & 3.83 & 95.89 & 95.20  & 97.53 & 87.80 & 94.11\\
\hline
\multicolumn{7}{c}{\textit{SCM(Conv($1 \times 1$)+Conv($3 \times 3$))}}\\
\hline
SCM(Iteration 1) & 4.30 & 94.15  & 92.87 & 97.17 & 87.63 & 92.95\\
SCM(Iteration 2)  & 3.76 & 96.62  & 91.27 & 97.13 & 90.07 & 93.77\\
SCM(Iteration 3)  & 3.55 & \textbf{97.81} & 91.67  & \textbf{97.60} & 89.53 & 94.15\\
\bottomrule
\end{tabular}
\end{center}
\end{table}

\subsection{Ablation Studies on the Structured Context Mixer }
\label{section7.1}
In this experiment, we investigate the influence of the proposed Structured Context Mixer with different numbers of iteration during the message passing process and  different kernel sizes of ${\lambda _c}$, ${\varpi _c}$ and ${\eta _c}$ shown in Algorithm \ref{Alg:mean-field}. Tables \ref{tab:scm} and \ref{tab:scm-mouse} show the performance comparisons of different networks on the PDMB validation and DeepLabcut Mouse Pose test datasets, respectively.We use 3 types of convolutional kernels 
where Conv($3 \times 3$) + Conv($1 \times 1$) means that the gate ${a_c}$ related kernels use $3 \times 3$ convolution with stride 1 and padding 1 to keep the same spatial resolution, while the gate ${g_c}$ related kernels use $1 \times 1$ convolution with stride 1 and padding 1. We do not choose larger convolutional kernels because this will affect the efficiency of the whole network. As shown in Table \ref{tab:scm}, with $3 \times 3$ convolutions, we achieve the lowest RMSE of 3.07 pixels and the highest mean PCK score of 96.08$\%$. When the iteration number is set to 2, we observe that, compared with the networks with other two types of SCM, the mean PCK score increases from 93.23$\%$ and 93.77$\%$ to 94.58$\%$ respectively by adopting SCM with $3 \times 3$ convolutions. The same trend can be observed when the iteration is set to 3. Actually, larger convolutional kernels help us to increase the receptive field of the network, which enables SCM to focus on larger context regions.

With respect to the number of iterations in the phase of inferring hidden structured context features, we train the networks with 1, 2 and 3 iterations, respectively. The results on the PDMB validation set show that the performances of the networks are continuously improved with the number of iterations increasing, which demonstrates that the proposed SCM can adaptively learn the structured contextual information of each keypoint considering the differences between mouse body parts. However, on the DeepLabCut Mouse Pose test set, the performance becomes worse when the number of iterations is set to 3, as shown in Table \ref{tab:scm-mouse}. The main reason is that the complexity of the network will increase as the number of iterations increases, requiring more samples for training. Overall, Although the network using SCM (Conv($3 \times 3$)) with 3 iterations holds the lowest RMSE and the highest PCK score on the PDMB validation set, a higher number of iterations will consume more computational overhead. Therefore, in our implementation, we set the number of iterations to 2 on both two datasets. In particular, all the convolutions in the first and second iterations share the same weights to reduce the parameters of the network for faster implementation.

For some challenging mouse keypoints on the PDMB dataset, e.g., tail base, which is occluded frequently, we receive the PCK score of 91.67$\%$, which is 4.8$\%$ improvement compared to the baseline model. This experimentally confirms that adding SCM to explore the keypoint-specific contextual information helps to determine the positions of the occluded parts.

\subsection{Ablation Studies on the Cascaded Multi-level Supervision Module }
\label{section7.2}

\begin{table}[htb]
\begin{center}
\caption{Ablation experiments for the Cascaded Multi-level Supervision module on the PDMB validation dataset (RMSE(pixels) and PCK@0.2).}
\label{tab:cmls}
\begin{tabular}{p{2.6cm}p{0.5cm}p{0.5cm}p{0.5cm}p{0.5cm}p{0.5cm}p{0.5cm}}
\toprule
Methods\quad & RMSE & Snout &  Left ear & Right ear & Tail base & Mean\\
\midrule

Baseline  & 4.89 & 93.38  & 92.67 & 93.97 & 86.87 & 91.72\\

    MLS(All Cas1)  & \textbf{4.45} & 93.49  & 96.10 & \textbf{98.16} & \textbf{86.93} & 93.67\\
CMLS(Start Cas2 )  & 5.75 & 92.13  & \textbf{97.20} & 97.93 & 86.37 & 93.41\\
CMLS(Middle Cas2)  & 4.98 & 94.78 & 91.07  & 96.63 & 85.60 & 92.02\\
CMLS(Final Cas2)  & 5.39 & \textbf{97.08} & 94.0  & 97.8 & 86.73 & \textbf{93.90}\\
CMLS(All Cas2)  & 13.61 & 95.19 & 93.07  & 96.33 & 78.83 & 90.85\\
\bottomrule
\end{tabular}
\end{center}
\end{table}

In this subsection, we investigate the effect of the CMLS module on the prediction results. In our experiments, we design 5 structures using the baseline model. MLS (All Cas1) refers to the case where we use the MLS block as intermediate supervision at the end of the first and second Hourglass modules and our SCM, while CMLS (All Cas2) refers to a cascaded structure composed of 2 identical MLS blocks. For CMLS (Start Cas2), CMLS (Middle Cas2) and CMLS (Final Cas2), we design the cascaded structure at the end of the first Hourglass, the second Hourglass and SCM, respectively, while the MLS blocks are kept on the other two positions. As shown in Tables \ref{tab:cmls} and \ref{tab:cmls-mouse}, we observe that all the networks with the CMLS module except the last one can improve the PCK score of the baseline model. However, we do not witness significant decrease on the RMSE of all the mouse parts for almost all the networks. The reason is that multi-level supervision applied to the inner layers of the network would affect the localisation results of the challenging mouse parts. As we can see from Table \ref{tab:cmls}, CMLS (All Cas2) has the PCK score of 78.83$\%$  for tail base, which is the worst PCK score among all the networks. On the other hand, compared with the baseline model, CMLS (Start Cas2) and MLS (All Cas1) obtain 4.53$\%$ and 4.19$\%$ improvement respectively for left and right ears,  demonstrating that our Cascaded Multi-level Supervision module can refine the prediction results providing large-scale (high-resolution) learning information. Furthermore, MLS (All Cas1) achieves the mean PCK score of 93.67$\%$ and the score is further improved to 93.90$\%$ by a designed cascaded structure applied to the final stage of SCM. This indicates that our CMLS module outperforms the single-scale supervision scheme.

\begin{table}
\begin{center}
\caption{Ablation experiments for the Multi-level Keypoints Aggregation on the PDMB validation dataset (RMSE(pixels) and PCK@0.2).}
\label{tab:postprocessing}
\begin{tabular}{p{3.0cm}p{0.5cm}p{0.5cm}p{0.5cm}p{0.5cm}p{0.5cm}p{0.5cm}}
\toprule
Methods\quad & RMSE & Snout &  Left ear & Right ear & Tail base & Mean\\
\midrule

SCM+CMLS(w/o MLKA)  & 2.89 & 97.35  & 98.70 & \textbf{98.80} & 90.90 & 96.44\\

SCM+CMLS(with small-scale MLKA)  & 2.90 & \textbf{97.84}  & 99.17 & 98.60 & \textbf{91.87} & 96.87\\
SCM+CMLS(with large-scale MLKA)   & \textbf{2.81} & 97.60  & 98.90 & 98.67  & 91.40 & 96.64\\
SCM+CMLS(with multi-scale MLKA)   & \textbf{2.81} & \textbf{97.84} & \textbf{99.20}  & 98.60 & \textbf{91.87} & \textbf{96.88}\\
\bottomrule
\end{tabular}
\end{center}
\end{table}

\begin{figure*}[htbp]
\begin{center}
\includegraphics[width=16cm]{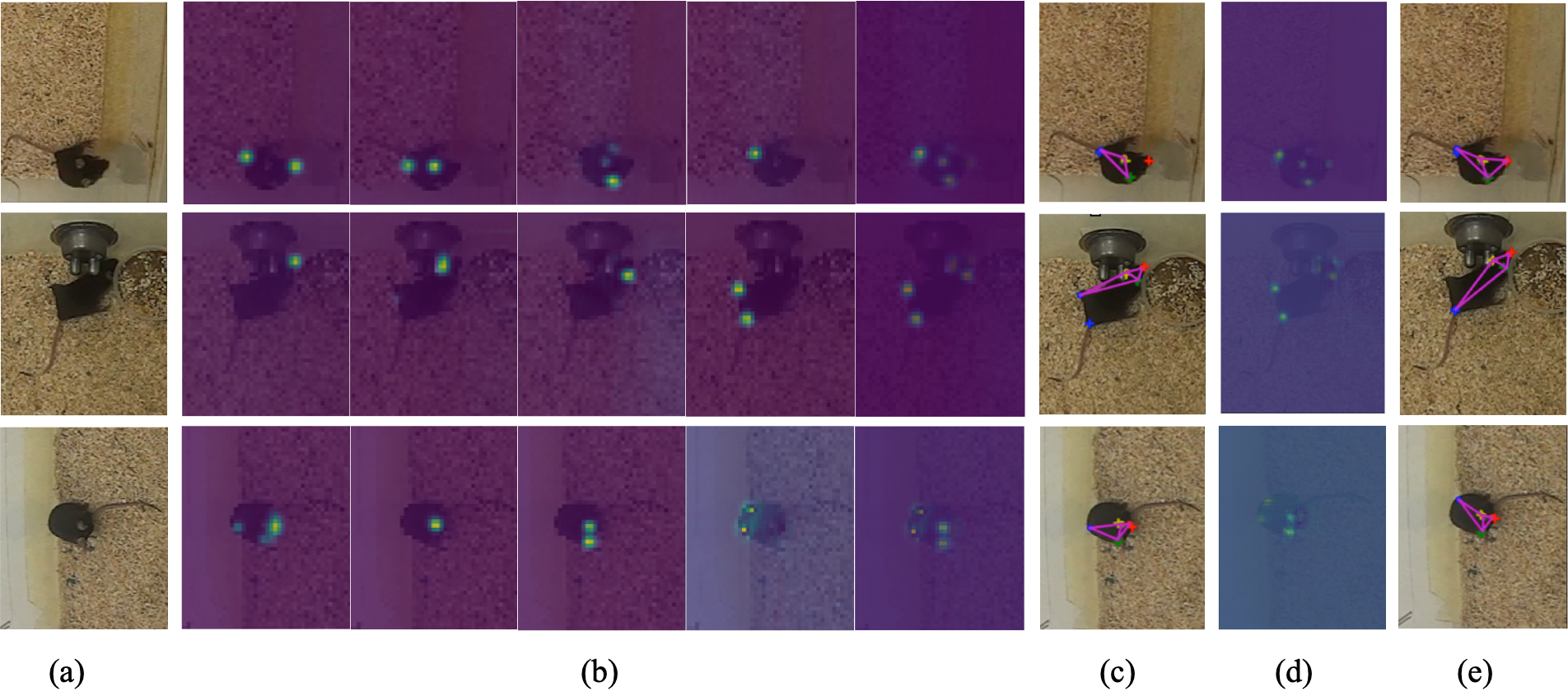}
\end{center}
\caption{{Visualisation of the prediction results from the Structured Context Mixer and the Cascaded Multi-level Supervision module of our GM-SCENet. (a) The input images (we crop the original images (640*480) for better display). (b) Heatmaps predicted by the Structured Context Mixer for 4 parts (160*120). The last column in (b) are the heatmaps of all the body parts. (c) Mouse skeletons generated by the Structured Context Mixer (Dots with different colors represent the predicted results of different parts, the symbol ‘+’ represents the ground truth, and the purple lines are the skeletons). (d) Further refined heatmaps (320*240) by the Cascaded Multi-level Supervision module. (e) Further refined skeletons by the Cascaded Multi-level Supervision module. } }
\label{fig:qualitative_SCM_CMLS }
\end{figure*}

\subsection{Ablation Studies on Multi-level Keypoints Aggregation}
\label{section7.3}
The CMLS module generates multi-level features for learning during the training phase, as described in Section \ref{section3.2}. After having obtained all the multi-level predicted keypoints, we aggregate all the chosen keypoints via the MLKA strategy. In our experiments, we design 3 schemes to evaluate the performance of the proposed MLKA where we use the network consisting of the 3-stack Hourglass network, SCM (Conv($3 \times 3$), Iteration 2) and CMLS (Final Cas2) for comparison. IN the first scheme, we generate keypoints by aggregating all the selected small-scale keypoints and the predicted keypoints from the last stage. In the other two schemes, we aggregate all the selected large- and multi-scale keypoints respectively. The threshold of the keypoint similarity in all the experiments is set to 0.2. As shown in Table \ref{tab:postprocessing}, we observe that the inference strategy combined with small-, large- and multi-scale MLKA respectively can improve the mean PCK score. In addition, the inference strategy with multi-scale MLKA has the lowest RMSE and the best mean PCK score among these three schemes on the PDMB validation set. Besides, we have 0.24$\%$ and 0.01$\%$ improvement by replacing the small- and large-scale MLKA strategies with the multi-scale MLKA respectively, and obtain 0.44$\%$ improvement by applying multi-scale MLKA to the inference phase of the baseline network.

\subsection{Comprehensive Analysis on Structured Context Mixer and  Cascaded Multi-level Supervision Module}
\label{section7.4}
We also investigate the contribution of the proposed Structured Context Mixer and the Cascaded Multi-level Supervision module to the whole network on both PDMB validation and DeepLabCut Mouse Pose test sets. In addition to adding each proposed module to the baseline model separately, we further employ the two proposed modules to the baseline model simultaneously. The results are reported in Table \ref{tab:ablationstudy}. On the two datasets, our method achieves the highest mean PCk scores, 96.44$\%$ and 97.89$\%$ respectively, with the two proposed modules onboard simultaneously, which are of 4.72$\%$ and 7.15$\%$ increment compared to the baseline model. Additionally, the network combining  SCM and the CMLS module achieves the lowest RMSE of 2.89 pixels on the PDMB validation set and the lowest RMSE of 3.13 pixels on the DeepLabCut Mouse Pose test set, respectively.

\begin{table}[htbp]
\begin{center}
\caption{Ablation experiments for the Structured Context Mixer and the Cascaded Multi-level Supervision module on the PDMB validation and DeepLabCut Mouse Pose test datasets (RMSE(pixels) and PCK@0.2).}
\label{tab:ablationstudy}
\begin{tabular}{p{1.4cm}p{0.5cm}p{0.5cm}p{0.5cm}p{0.5cm}p{0.5cm}p{0.5cm}p{0.5cm}p{0.5cm}}
\toprule
Dataset &SCM &CMLS & RMSE & Snout &  Left ear & Right ear & Tail base & Mean\\
\midrule

\multirow{4}*{PDMB}&  &  & 4.89 & 93.38  & 92.67 & 93.97 & 86.87 & 91.72\\
&$\surd$  &   & 3.74 & 96.90  & 96.50 & 96.07 & 88.87 & 94.58\\
&  & $\surd$  & 5.39 & 97.08 & 94.0  & 97.8 & 86.73 & 93.90\\
&$\surd$  & $\surd$ & \textbf{2.89} & \textbf{97.35}  & \textbf{98.70} & \textbf{98.80} & \textbf{90.90} & \textbf{96.44}\\
\midrule
\multirow{4}{1.4cm}{DeepLabCut Mouse Pose}&  &  & 3.83 & 89.35 & 90.28 & 90.28  & 93.06 & 90.74\\
&$\surd$  &  & 3.67  & 95.77 & 96.24 & 90.61 & 98.12 & 95.19\\
&  & $\surd$ & 3.52  & 94.84 & 97.65 & 95.77 & 97.65 & 96.48\\
&$\surd$  & $\surd$ & \textbf{3.13} & \textbf{96.71} & \textbf{98.60}  & \textbf{97.65} & \textbf{98.59} & \textbf{97.89}\\

\bottomrule
\end{tabular}
\end{center}
\end{table}

 \begin{table*}[htbp]
\begin{center}
\caption{Comparisons of RMSE and PCK@0.2 score on the PDMB and DeepLabCut Mouse Pose test sets.}
\label{tab:comparision}
\begin{tabular}{p{3.6cm}|p{0.6cm}p{0.6cm}p{0.6cm}p{0.6cm}p{0.6cm}p{0.6cm}|p{0.6cm}p{0.6cm}p{0.6cm}p{0.6cm}p{0.6cm}p{0.6cm}}
\toprule
Dataset &
\multicolumn{6}{c|}{PDMB} &
\multicolumn{6}{c}{DeepLabCut Mouse Pose}\\
\midrule
Methods\quad & RMSE & Snout &  Left ear & Right ear & Tail base & Mean & RMSE & Snout &  Left ear & Right ear & Tail base & Mean\\
\midrule
\midrule
Newell et al.\cite{newell2016stacked}   & 4.29 & 95.69  & 92.75 & 95.05 & 93.38 & 94.22 &5.80 & 92.96 & 93.43 & 91.55 & 88.73 & 91.67\\

Mathis et al.\cite{Mathis2018}  & 3.73 & 96.50  & 98.25 & 97.85 & 93.58 & 96.54 & 3.64 &93.90  &93.90 & 85.92 & \textbf{98.59} & 93.08\\
Pereira et al.\cite{Pereira2019}  & 4.0 & 94.68  & 94.10 & 96.25 & 93.23 & 94.56  & 5.70 & 88.26 & 94.37 & 94.37 & 96.24 & 93.31\\
Graving et al.\cite{Graving2019}   & 3.23 & 96.96 & 98.89  & 97.30 & 94.64 & 96.95 & 3.48 & 96.24 & 95.31 & 92.02 & 98.12 & 95.42\\
Toshev et al.\cite{toshev2014deeppose}  & 5.78 & 83.92 & 83.75  & 89.90 & 63.21 & 80.20 & 7.30 & 73.24 & 64.32 & 72.30 & 79.34 & 72.30\\
Wei at al.\cite{Wei2016}  & 4.71 & 93.41 & 92.05  & 92.60 & 94.69 & 93.19 & 6.19 & 91.08 & 89.67 & 81.22 & 96.71 & 89.67\\
Chu et al.\cite{chu2017multi}  & 4.20 & 96.65  & 94.35 & 97.0 & 94.64 & 95.66  &5.54 & 91.08 & 92.49 & 91.55 & 96.71 & 92.96\\
Sun et al.\cite{Sun2019}  & 3.91 & 96.50  &97.0 & 97.25 & 95.83 & 96.65  &4.66 & 90.61 & 91.08 & 94.84 & 98.12 & 93.66\\
Artacho et al.\cite{artacho2020unipose}  & 4.37 & 95.38 & 97.25 & 97.10 & 93.13 & 95.72  &3.87 & 95.31 & 94.84 & 94.84 & 95.77 & 95.19\\
Ours(SCM)  & 3.46 & 96.91 & 97.60  & 98.20 & 95.19 & 96.97 & 3.67 & 95.77 & 96.24 & 90.61 & 98.12 & 95.19\\
Ours(CMLS)  & 4.79 & 96.45 & 95.50  & 97.30 & 93.13 & 95.60 & 3.52  & 94.84 & 97.65 & 95.77 & 97.65 & 96.48 \\
Ours(GM-SCENet) & 2.84 & \textbf{97.87} & 98.65  & 98.80 & 95.89 & 97.80 &3.13 & 96.71 & \textbf{98.60}  & 97.65 & \textbf{98.59} & 97.89\\
Ours(GM-SCENet+MLKA)  & \textbf{2.79} & 97.72 & \textbf{98.90}  & \textbf{99.10} & \textbf{96.34} & \textbf{98.01} & \textbf{3.10} & \textbf{97.65} & 98.12 & \textbf{98.12} & \textbf{98.59} & \textbf{98.12}\\
\bottomrule
\end{tabular}
\end{center}
\end{table*}

 We also observe that the overall performance of the network combined with only the CMLS module outperforms that of the SCM based network on the DeepLabCut Mouse Pose dataset. Different from the tail base in the PDMB dataset, the CMLS based network does not result in much decrease on the PCK score of snout which is a relatively challenging part in the DeepLabCut Mouse Pose dataset. On the contrary, the CMLS based network improves the PCK score of snout from 89.35$\%$ to 94.84$\%$. As we discussed in Section \ref{section7.2}, the localisation errors of the challenging mouse parts would be amplified when only the CMLS module is used, although those of relatively easy parts can be further reduced. However, the CMLS based network can effectively deal with the problem of scaling variations by providing multi-level supervised information. Actually, the size of the mouse in the input image is different since the images in the DeepLabCut Mouse Pose dataset have different resolutions. Therefore, our network can achieve better performance by combining both SCM and the CMLS module, where SCM can adaptively learn and enhance structured contextual information by exploring the differences between different parts and the CMLS can provide multi-level supervised information to strengthen the contextual feature learning.

 Finally, we explore the effect of the CMLS module applied to the last stage of SCM using qualitative evaluation on our PDMB validation set, as demonstrated in Fig. \ref{fig:qualitative_SCM_CMLS }. We observe that the network leads to more accurate localisation results by adding the CMLS module to SCM in the challenging cases such as abnormal mouse poses, occlusion and invisibility. The network without the CMLS module produces the failing predictions as shown in Fig. \ref{fig:qualitative_SCM_CMLS }(c). For example, in the 2nd row of Fig. \ref{fig:qualitative_SCM_CMLS }(c), the mouse's left foot was mispredicted as the tail base due to the highly deformable body. However, adding the CMLS module to SCM results in the refined predictions as shown in Fig. \ref{fig:qualitative_SCM_CMLS }(e). It is also noticeable that the improved network produces higher responses in the ambiguous regions of the heatmaps, as depicted in Fig. \ref{fig:qualitative_SCM_CMLS }(d), compared to the 5th column of Fig .\ref{fig:qualitative_SCM_CMLS }(b). This difference is because the last CMLS module can preserve the spatial correlations between different parts, where the predicted heatmaps in SCM can be used as prior knowledge to combine multi-level features from the baseline model.

\subsection{Comparison with State-of-the-art Pose Estimation Networks}
\label{section7.5}
In this section, we report the RMSE and PCK scores of our proposed and the other state-of-the-art methods with the threshold of 0.2 on our PDMB and DeepLabCut Mouse Pose test sets, as shown in Table \ref{tab:comparision}. Among them, there are three state-of-the-art approaches of animal pose estimation and six networks of human pose estimation.

\begin{figure}[htbp]
\begin{center}
\begin{tabular}{cc}
\includegraphics[width=4.8cm]{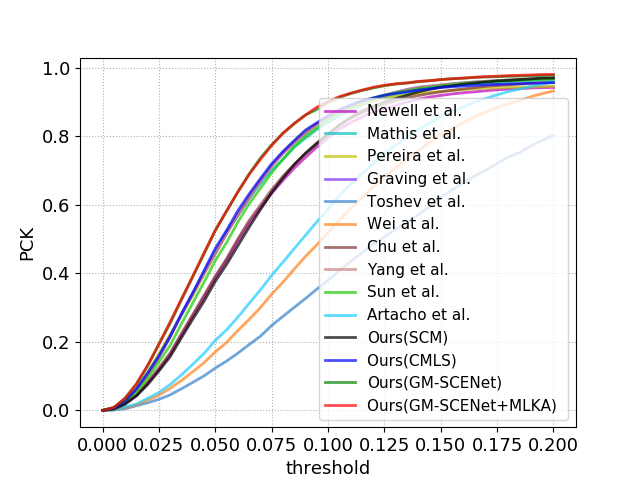}&
\hspace{-8mm}
\includegraphics[width=4.8cm]{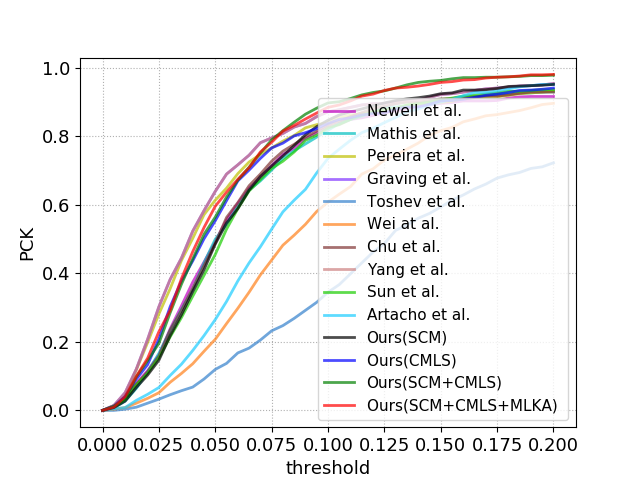}\\
\hspace{-8mm}
(a) PDMB & (b) DeepLabCut
\end{tabular}
\end{center}
\caption{Comparisons of mean PCK curves on the PDMB and DeepLabCut Mouse Pose test sets.}
\label{fig:pdmousemeanpckcurve}
\end{figure}

On the PDMB test set, our proposed GM-SCENet achieves the state of the art RMSE of 2.84 pixels and the mean PCK score of 97.80$\%$. Compared with the Stacked DenseNet for animal pose estimation \cite{Graving2019}, our GM-SCENet improves the mean PCK score from 96.95$\%$ to 97.80$\%$, and the performance can be further improved to 98.10$\%$ by adding the MLKA inference algorithm. In particular, for the most challenging mouse part, i.e., tail base, our GM-SCENet achieves 1.25$\%$ improvement. The PCK scores for snout, left and right ears are also improved to 97.87$\%$, 98.65$\%$ and 98.80$\%$, respectively. Among these popular networks for human pose estimation, the model of \cite{toshev2014deeppose} has the worst performance. This is because it directly predicts the numerical value of each coordinate rather than the heatmap of each part by a CNN. Also, their predictions ignore the spatial relationships between keypoints in an image. Compared with the 8-stack Hourglass network of \cite{newell2016stacked}, our GM-SCENet (3-stack Hourglass network based) outperforms it by a large margin, where the mean PCK score holds 3.58$\%$ improvement. The HRNet \cite{Sun2019} achieves the best performance among the six human pose estimation methods. Nevertheless, we observe that our SCM based network can achieve competitive performance in comparison with HRNet and other approaches, demonstrating the effectiveness of our proposed SCM for structured context enhancement. The performance can be further improved by adding the CMLS module. Figs. \ref{fig:pdmousemeanpckcurve}(a) and \ref{fig:pdmousepckcurve} show the mean PCK curves of different methods and the PCK curves for each keypoint, respectively. Our method (GM-SCENet+MLKA) achieves the best performance with different thresholds among all the methods. The final results demonstrate the superior performance of our proposed network over the other  state-of-the-art methods in terms of PCK@0.2 score.

On the DeepLabCut Mouse Pose test set, our SCM based network achieves the mean PCK score of 95.19$\%$, which exceeds the results of the other state-of-the-art networks except for the Stacked DenseNet. However, our CMLS based network obtains 1.06$\%$ improvement compared with the Stacked DenseNet. Furthermore, the proposed GM-SCENet improves the PCK scores with large margins of 3.29$\%$ and 2.81$\%$ on the left and right ears, compared with the closest competitors, and achieves 2.47$\%$ improvement in average, compared with the Stacked DenseNet. The use of the MLKA scheme results in further improvement on the mean PCK score and decreases on the keypoint RMSE, where our method results in an improvement of 2.93$\%$ over the best result of the methods used for human pose estimation, i.e., Unipose\cite{artacho2020unipose}. Figs. \ref{fig:pdmousemeanpckcurve}(b) and \ref{fig:mousepckcurve} show the mean PCK curves and the PCK curves of different methods for individual parts. The GM-SCENet with or without MLKA can achieve competitive performance, compared to the other state-of-the-art methods, on the DeepLabCut Mouse Pose test set.

It is also noteworthy that although our PDMB dataset contains more challenging mouse poses, the performance of different methods on this dataset outperforms that on the DeepLabCut Mouse Pose dataset. The reason is that our PDMB is a large database which has more images for training. For the Small DeepLabCut Mouse Pose dataset, our proposed method can achieve significant improvement on the PCK score, compared with the other state-of-the-art approaches. This shows that our method works satisfactorily on both small and relatively large datasets. Fig. \ref{fig:mousevis} gives some examples of the estimated mouse poses on the PDMB and DeepLabCut test sets.

\begin{figure}
\begin{center}
\begin{tabular}{cc}
\includegraphics[width=4.8cm]{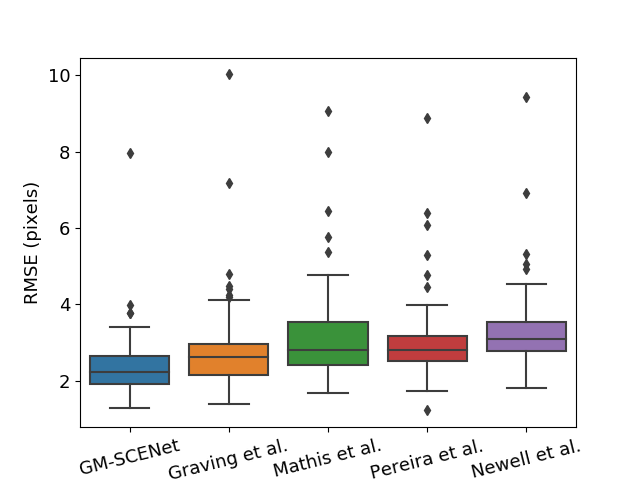}&
\hspace{-9mm}
\includegraphics[width=4.8cm]{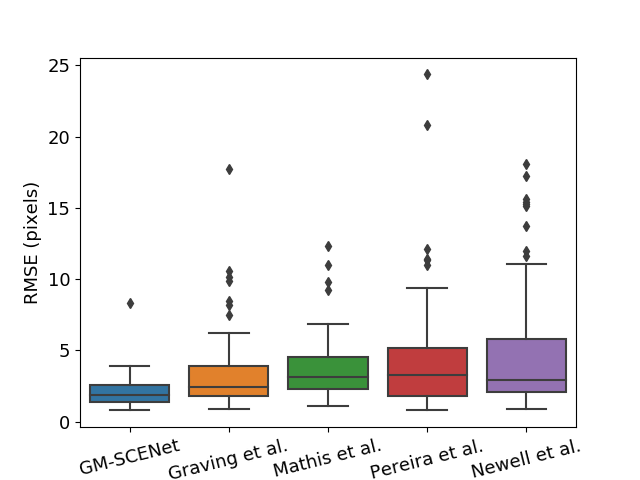}\\
\hspace{-9mm}
(a) PDMB  &(b) DeepLabCut
\end{tabular}
\end{center}
\caption{Comparisons of keypoints' RMSE on the PDMB and DeepLabCut test sets. The rectangles corresponding to individual methods are used to depict the distribution of the keypoints' RMSE with the minimum and maximum values.}
\label{fig:rmsebox}
\end{figure}

We also compare the distribution of keypoint RMSE on the PDMB test and DeepLabCut test sets (we randomly choose a group of samples rather than using the whole test set for comparison in our experiments). Here, the box plots are plotted in Fig. \ref{fig:rmsebox}(a) and (b) to measure the effectiveness of our proposed method and other four comparison methods on the two datasets in terms of keypoints' RMSE and stability. Fig \ref{fig:rmsebox}(a) and (b) show that our GM-SCENet sustains the lowest keypoints' RMSE with the lowest variance on the two datasets. This demonstrates that our proposed GM-SCENet can reduce the keypoints' RMSE effectively and produce more stable pose estimation results.

To further verify the generalisation capability of our proposed methods, we also conduct experiments on the Zebra, Fly and LSP datasets (Supplementary D, F). The results reported in Table \ref{tab:zebra}, \ref{tab:fly} show that our GM-SCENet can achieve competitive performance on Fly and Zebra in which the objects own a body structure similar to that of mice in comparison to the other state-of-the-art approaches. Additionally, Table \ref{tab:human} exhibits good generalisation of our model on human pose domain. Fig. \ref{fig:zebravis} illustrates some examples of the estimated poses on the Zebra and Fly datasets.

\section{Conclusion}
We have presented a novel end-to-end architecture called Graphical Model based Structured context Enhancement Network (GM-SCENet) for mouse pose estimation. By concentrating on the differences between different mouse body parts, the proposed Structured Context Mixer (SCM) based on a novel graphical model is able to adaptively learn the structured context information of individual mouse parts, including global and keypoint-specific contextual features. Meanwhile, a Cascaded Multi-level Supervision module (CMLS) has been designed to jointly train SCM and the backbone network by providing multi-level supervised information, which strengthens contextual feature learning and improves the robustness of our network. We also designed an inference strategy, i.e., Multi-level Keypoint Aggregation (MLKA), where the selected multi-level keypoints are aggregated for better prediction. Experimental results on our Parkinson’s Disease Mouse Behaviour (PDMB) and the standard DeepLabCut Mouse Pose datasets demonstrate that the proposed approach outperforms several baseline methods. Additionally, our method can also be leveraged to estimate mouse pose with more keypoints and potentially extended to side-view or other datasets, especially for pose estimation of small objects with highly deformable body structures. Our future work will explore social behaviour analysis of mice with Parkinson’s Disease using the proposed mouse pose estimation model.


\ifCLASSOPTIONcaptionsoff
  \newpage
\fi

\bibliographystyle{IEEEtran}
\bibliography{mybibfile}
\vspace{-12 mm}

\begin{IEEEbiographynophoto}{Feixiang Zhou} is currently pursuing the PhD degree with the School of Informatics, University of Leicester, U.K.
\end{IEEEbiographynophoto}
\vspace{-12 mm}

\begin{IEEEbiographynophoto}{Zheheng Jiang} is currently a senior research associate at the School of Computing and Communications, Lancaster University, U.K.
\end{IEEEbiographynophoto}
\vspace{-12 mm}

\begin{IEEEbiographynophoto}{Zhihua Liu} is currently pursuing the PhD degree with the School of Informatics, University of Leicester, U.K.
\end{IEEEbiographynophoto}
\vspace{-12 mm}

\begin{IEEEbiographynophoto}{Fang Chen} is currently pursuing the PhD degree with the Department of Mathematics, University of Leicester, U.K.
\end{IEEEbiographynophoto}
\vspace{-12 mm}

\begin{IEEEbiographynophoto}{Long Chen} is currently pursuing the PhD degree with the Department of Informatics, University of Leicester, U.K.
\end{IEEEbiographynophoto}
\vspace{-12 mm}

\begin{IEEEbiographynophoto}{Lei Tong} is currently pursuing the PhD degree with the Department of Informatics, University of Leicester, U.K.
\end{IEEEbiographynophoto}
\vspace{-12 mm}

\begin{IEEEbiographynophoto}{Zhile Yang} is currently an associate professor in Shenzhen Institute of Advanced Technology, Chinese Academy of Sciences, Shenzhen, China.
\end{IEEEbiographynophoto}
\vspace{-12 mm}

\begin{IEEEbiographynophoto}{Haikuan Wang} is currently an associate professor at School of Mechanical Engineering and Automation, Shanghai University, Shanghai, China.
\end{IEEEbiographynophoto}
\vspace{-12 mm}

\begin{IEEEbiographynophoto}{Minrui Fei} is currently a professor and dean at School of Mechanical Engineering and Automation, Shanghai University, Shanghai, China.
\end{IEEEbiographynophoto}
\vspace{-12 mm}

\begin{IEEEbiographynophoto}{Ling Li} is currently a senior lecturer at School of Computing, University of Kent, U.K.
\end{IEEEbiographynophoto}
\vspace{-10 mm}

\begin{IEEEbiographynophoto}{Huiyu Zhou} currently is a Professor at School of Informatics, University of Leicester, U.K.
\end{IEEEbiographynophoto}
\vspace{-10 mm}

\clearpage
\onecolumn

\setcounter{table}{0}
\setcounter{algorithm}{0}
\setcounter{figure}{0}
\setcounter{equation}{0}
\setcounter{page}{1}
\renewcommand\thefigure{S\arabic{figure}}
\renewcommand\thetable{S\arabic{table}}
\renewcommand\thealgorithm{S\arabic{algorithm}}
\section*{Supplementary A}

\begin{table}[htbp]
\begin{center}
\caption{Ablation experiments of the Structured Context Mixer on DeepLabCut Mouse Pose test dataset (RMSE(pixels) and PCK@0.2).}
\label{tab:scm-mouse}
\begin{tabular}{p{2.6cm}p{1cm}p{1cm}p{1cm}p{1cm}p{1cm}p{1cm}}
\toprule
Methods\quad & RMSE & Snout &  Left ear & Right ear & Tail base & Mean\\
\midrule

Baseline  & 3.83 & 89.35 & 90.28 & 90.28  & 93.06 & 90.74\\
\hline

\multicolumn{7}{c}{\textit{SCM(Conv($3 \times 3$))}}\\
\hline
SCM(Iteration 1)  & 4.70 & 91.20  & 95.83 & 95.37 & 96.30 & 94.68\\
SCM(Iteration 2)  & 4.11 & 91.08  & \textbf{97.18} & \textbf{96.24} & \textbf{98.12} & \textbf{95.66}\\
SCM(Iteration 3)  & 14.87 & 91.08 & 76.53  & 69.95 & 88.26 & 81.46\\
\hline
\multicolumn{7}{c}{\textit{SCM(Conv($1 \times 1$))}}\\
\hline
SCM(Iteration 1) & 4.09 & 94.37  & 93.43 & 92.96 & 94.84 & 93.90\\
SCM(Iteration 2)  & \textbf{3.67} & \textbf{95.77}  & 96.24 & 90.61 & \textbf{98.12} & 95.19\\
SCM(Iteration 3)  & 4.52 & 92.49 & 90.61  & 93.90 & 94.84 & 92.96\\
\hline
\multicolumn{7}{c}{\textit{SCM(Conv($1 \times 1$)+Conv($3 \times 3$))}}\\
\hline
SCM(Iteration 1) & 3.97 & 95.31  & 97.65 & 88.26 & 95.30 & 94.13\\
SCM(Iteration 2)  & 3.93 & 95.77  & 96.71 & 89.67 & 95.31 & 94.37\\
SCM(Iteration 3)  & 8.55 & 91.08 & 89.67 & 84.04 & 81.69 & 86.62\\
\bottomrule
\end{tabular}
\end{center}
\end{table}

\begin{table}[htb]
\begin{center}
\caption{Ablation experiments of the Cascaded Multi-level Supervision module on the DeepLabCut Mouse Pose test set (RMSE(pixels) and PCK@0.2).}
\label{tab:cmls-mouse}
\begin{tabular}{p{2.6cm}p{1cm}p{1cm}p{1cm}p{1cm}p{1cm}p{1cm}}
\toprule
Methods\quad & RMSE & Snout &  Left ear & Right ear & Tail base & Mean\\
\midrule

Baseline  & 3.83 & 89.35 & 90.28 & 90.28  & 93.06 & 90.74\\

MLS(All Cas1)  & 7.85 & 90.28  & 91.67 & 95.37 & \textbf{98.61} & 93.98\\
CMLS(Start Cas2 )  & \textbf{3.52}  & \textbf{94.84} & \textbf{97.65} & \textbf{95.77} & 97.65 & \textbf{96.48}\\
CMLS(Middle Cas2)  & 5.35 & 90.14 & 90.61  & 88.26 & 98.59 & 91.90\\
CMLS(Final Cas2)  & 4.51 & 89.20 & 93.42  & 95.77 & 98.12 & 94.13\\
CMLS(All Cas2)  & 8.01 & 81.22 & 93.43  & 92.02 & 96.24 & 90.73\\
\bottomrule
\end{tabular}
\end{center}
\end{table}

\clearpage
\onecolumn
\section*{Supplementary B}

\begin{figure}[htb]
\begin{center}
\begin{tabular}{cc}
\includegraphics[width=8.8cm]{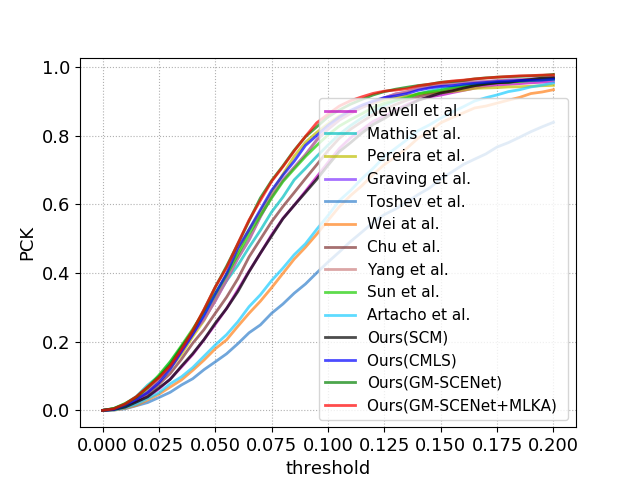}&
\hspace{-8mm}
\includegraphics[width=8.8cm]{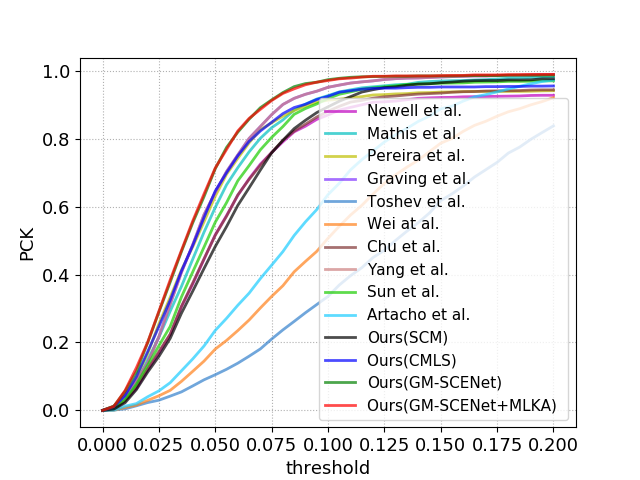}\\
\hspace{-8mm}
(a) Snout &(b) Left ear\\
\includegraphics[width=8.8cm]{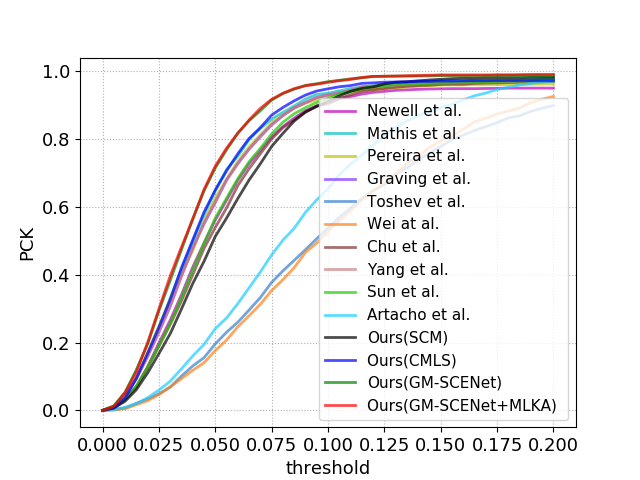}&
\hspace{-8mm}
\includegraphics[width=8.8cm]{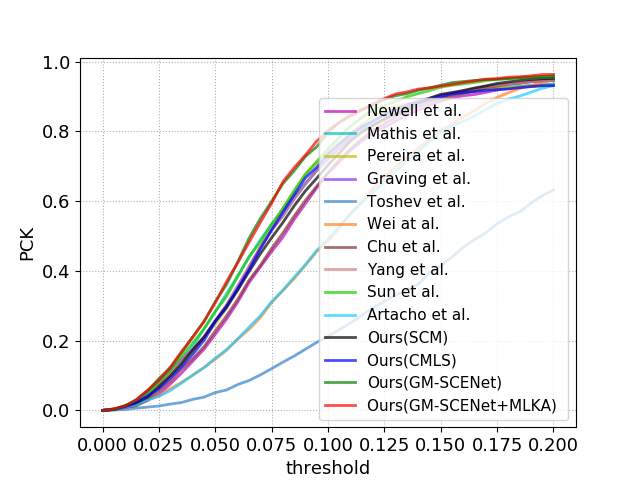}\\
\hspace{-8mm}
 (c) Right ear &(d) Tail base
\end{tabular}
\end{center}
\caption{Comparisons of the PCK curves for each part on the PDMB test set.}
\label{fig:pdmousepckcurve}
\end{figure}

\clearpage
\onecolumn
\section*{Supplementary C}

\begin{figure}[htbp]
\begin{center}
\begin{tabular}{cc}
\includegraphics[width=8.8cm]{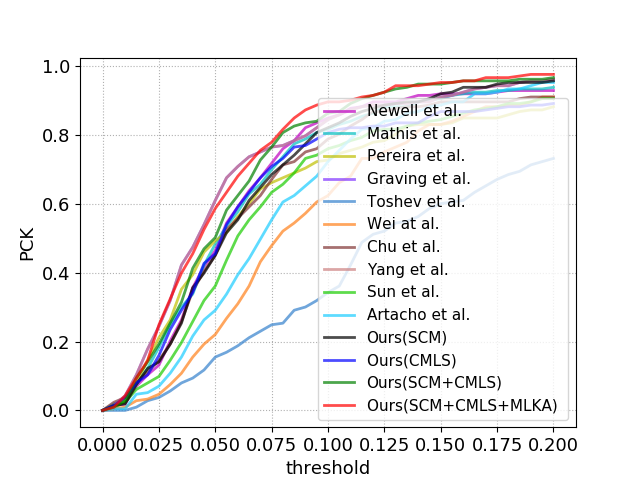}&
\hspace{-8mm}
\includegraphics[width=8.8cm]{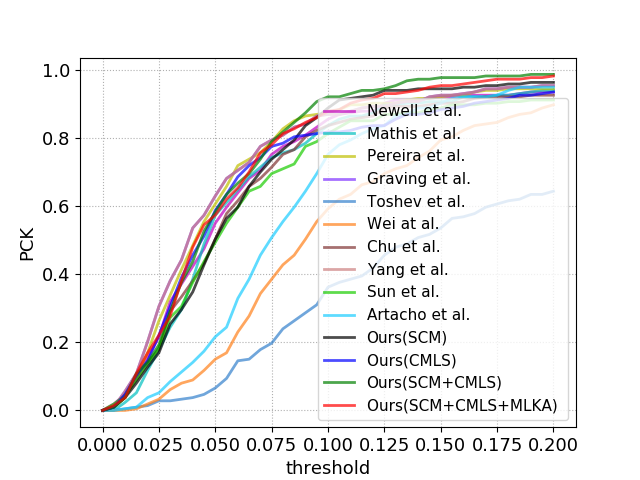}\\
\hspace{-8mm}
(a) Snout &(b) Left ear \\
\includegraphics[width=8.8cm]{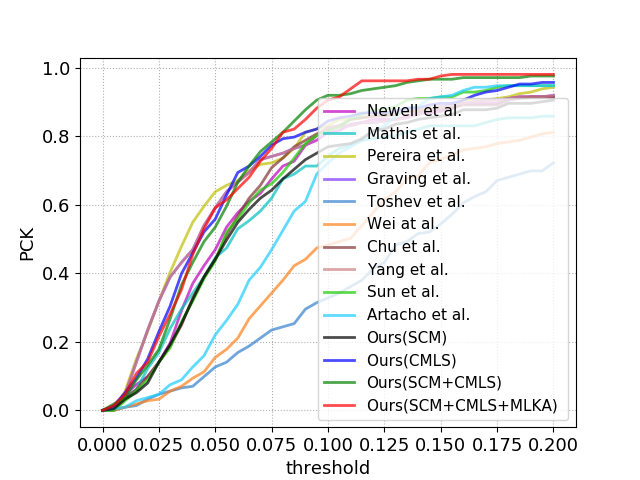}&
\hspace{-8mm}
\includegraphics[width=8.8cm]{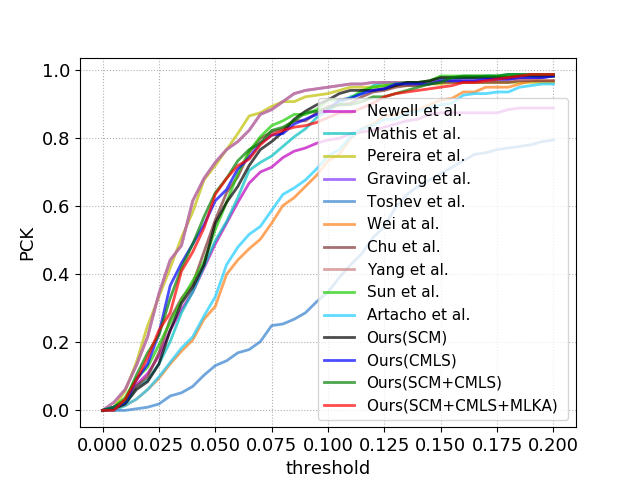}\\
\hspace{-8mm}
(c) Right ear &(d) Tail base
\end{tabular}
\end{center}
\caption{Comparisons of the PCK curves for each part on the DeepLabCut Mouse Pose test set.}
\label{fig:mousepckcurve}
\end{figure}

\clearpage
\onecolumn
\section*{Supplementary D}

\begin{table*}[htbp]
\begin{center}
\caption{Comparisons of RMSE and PCK@0.2 score on the zebra test set.}
\label{tab:zebra}
\begin{tabular}{p{3.6cm}p{0.5cm}p{0.5cm}p{0.5cm}p{0.5cm}p{0.9cm}p{0.9cm}p{0.9cm}p{0.9cm}p{0.5cm}p{0.5cm}p{0.5cm}}
\toprule
Methods\quad & RMSE & Snout &  Head & Neck & ForelegL1 & ForelegR1 & HindlegL1 & HindlegR1 & Tail base & Tail tip & Mean\\
\midrule

Newell et al.\cite{newell2016stacked}   & 2.09 & 81.67  & 85.0 & 98.33 & 96.67 & 92.78 &97.22 & 97.22 & 97.78 & 76.67 & 91.48\\

Mathis et al.\cite{Mathis2018}  & 1.71 & 81.67 & 90.0 & 100.0 & 97.22 & 97.22 & 97.78 &98.33  &\textbf{100.0} & 76.67 & 93.21 \\
Pereira et al.\cite{Pereira2019}  & 4.13 & 72.22 & 80.56 & 90.56 & 91.11  & 94.44 & 93.33 & 92.22 & 93.33 & 80.56 &87.59 \\
Graving et al.\cite{Graving2019}   & 1.62 & 83.89 & 92.78  & \textbf{100.0} & \textbf{97.78} & 96.67 & \textbf{98.89} & \textbf{98.89} & \textbf{100.0} & 77.22 & 94.01\\
Toshev et al.\cite{toshev2014deeppose}  & 2.35 & 77.22  & 65.0 & 97.78 & 95.0 & 91.11 & 91.67 & 91.11 & 98.33 &72.78 & 86.67\\
Wei at al.\cite{Wei2016}  & 3.00 & 66.67 & 71.11  & 95.0 & 55.56 & 65.56 & 57.22 & 65.56 & 95.56 & 60.0 & 70.25\\
Chu et al.\cite{chu2017multi}  & 1.87 & 85.56  & 85.0 & 98.33 & 96.67 & 95.0  &96.67 & 98.33 & 98.89 & 82.22 & 92.96\\
Sun et al.\cite{Sun2019}  & 2.91 & 81.11  &90.0 & 97.78 & 95.56 & 93.33  &93.89 & 95.56 & 95.0 & 74.44 & 90.74 \\
Artacho et al.\cite{artacho2020unipose}  & 2.90 & 71.11 & 73.89 & 91.11 & 58.33 & 68.33  &59.44 & 62.78 & 96.11 & 65.56 & 71.85\\

Ours(SCM)  & 1.85 & 83.33 & 93.33  & 96.67 & 97.22 & 97.22 & 97.22 & 98.33 & 98.33 & 83.89 & 93.95\\
Ours(CMLS)  & 1.94 & 83.33 & 86.11  & 99.44 & 97.22 & 96.67 & 97.22  & 98.33 & 98.33 & 75.56 & 92.47 \\
Ours(GM-SCENet) & 1.60 & 87.22 & 93.33  & \textbf{100.0} & 97.22 & \textbf{97.78} &99.44 & 97.78 & 99.44  & \textbf{86.11} & 95.37\\
Ours(GM-SCENet+MLKA)  & \textbf{1.57} & \textbf{88.33} & \textbf{93.89}  & 99.44 & \textbf{97.78} & \textbf{97.78} & \textbf{98.89} & 98.33 & 99.44 & 85.56 & \textbf{95.49}\\
\bottomrule
\end{tabular}
\end{center}
\end{table*}

\begin{table*}[htbp]
\begin{center}
\caption{Comparisons of RMSE and PCK@0.2 score on the fly test set. We choose 8 difficult keypoints of the fly, i.e.,MidlegR3, HindlegR2, HindlegR3, HindlegR4, MidlegL4, HindlegL2, HindlegL3, HindlegL4  for comparison.}
\label{tab:fly}
\begin{tabular}{p{3.6cm}p{0.5cm}p{0.9cm}p{0.9cm}p{0.9cm}p{0.9cm}p{0.9cm}p{0.9cm}p{0.9cm}p{0.9cm}p{0.9cm}}
\toprule
Methods\quad & RMSE & MidlegR3  & HindlegR2 & HindlegR3 & HindlegR4 & MidlegL4 & HindlegL2 & HindlegL3 & HindlegL4& Mean\\
\midrule

Newell et al.\cite{newell2016stacked}   & 2.88 & 96.67  & 90.0 & 85.67 & 87.0 &95.0 & 93.67 & 89.0 & 87.67 & 90.58\\

Mathis et al.\cite{Mathis2018}  & 1.94 & 96.67 & 93.0 & 92.0 & 83.67 & 93.67 &94.0  &92.0 & 88.0 & 91.63 \\
Pereira et al.\cite{Pereira2019}  & 2.99 & 94.67 & 89.67 & 86.0  & 82.33 & 91.67 & 91.0 & 89.0 & 86.33 &88.83 \\
Graving et al.\cite{Graving2019}   & \textbf{1.91} & 96.67  & 92.67 & 92.0 & 85.67 & 95.33 & 94.0 & 92.33 & 87.0 & 91.96\\
Toshev et al.\cite{toshev2014deeppose}  & 3.26 & 90.33  & 87.33 & 85.0 & 59.0 & 89.33 & 89.33 & 84.33 &72.0 & 82.08\\
Wei at al.\cite{Wei2016}  & 3.03 & 93.67   & 89.33 & 88.67 & 80.67 & 94.0 & 89.0 & 88.0 & 83.33 & 88.33\\
Chu et al.\cite{chu2017multi}  & 2.59 & 95.67  & 90.0& 90.33 & 88.33  &91.67 & 90.33 & 87.0 & 91.0 & 90.54\\

Sun et al.\cite{Sun2019}  & 7.43   &55.0 & 91.67 & 93.0 & 82.0  &94.0 & 92.67 & 92.33 & 87.33  &86.0\\
Artacho et al.\cite{artacho2020unipose}  & 1.99 & 98.0 & 94.0 & 88.67 & 82.33 &95.67 & 94.67 & 86.67 & 87.67 & 90.96\\

Ours(SCM)  & 2.40  & 98.33  & 92.67 & 91.67 & 88.33& 95.0 & 92.67 & 90.33 & 91.33 & 92.54\\
Ours(CMLS)  & 3.17 & 96.67 & 91.0 & 88.67 & 87.0 & 92.67  & 94.0 & 90.67 & 88.0 & 91.08\\
Ours(GM-SCENet) & 2.56 & 98.33  & \textbf{94.0} & \textbf{94.0} & \textbf{90.67} &\textbf{96.33} & \textbf{96.33} & \textbf{94.0}  & 94.33 & \textbf{94.75}\\
Ours(GM-SCENet+MLKA)  & 2.51 & \textbf{98.67}   & \textbf{94.0} & 93.33 & \textbf{90.67} & 95.67 & \textbf{96.33} &\textbf{94.0} & \textbf{94.67}& 94.67\\
\bottomrule
\end{tabular}
\end{center}
\end{table*}

\begin{table*}[htbp]
\begin{center}
\begin{threeparttable} 
\caption{Comparisons of PCK@0.2 score on the LSP test set.}
\label{tab:human}
\begin{tabular}{p{3.6cm}p{0.9cm}p{0.9cm}p{0.9cm}p{0.9cm}p{0.9cm}p{0.9cm}p{0.9cm}p{0.9cm}}
\toprule
Methods\quad  & Head  & Sho. & Elb. & Wri. & Hip & Knee & Ank.& Mean\\
\midrule

Lifshitz et al.*   & 96.8 & 89.0 & 82.7  &79.1 & 90.9 & 86.0 & 82.5 & 86.7\\
Wei at al.\cite{Wei2016}  & 97.8 & 92.5 & 87.0 & 83.9 & 91.5 & 90.8 & 89.9 & 90.5\\
Bulat at al.*  & 97.2 & 92.1 & 88.1 & 85.2 & 92.2 & 91.4 & 88.7 & 90.7\\
Chu et al.\cite{chu2017multi}   & 98.1 & 93.7 & 89.3  &86.9 & 93.4 & 94.0 & 92.5 & 92.6\\

Shu et al.*   & 97.9 & 93.6 & 89.0  &85.8 & 92.9 & 91.2 & 90.5 & 91.6\\
Yang et al.*   & 98.3 & 94.5 & 92.2  &88.9 & 94.4 & 95.0 & 93.7 & 93.9\\
Zhang et al.*   & 98.4 & 94.8 & 92.0  &89.4 & 94.4 & 94.8 & 93.8 & 94.0\\ 
Cao et al.*   & \textbf{98.6} & 95.1 & 92.1  &89.8 & 94.7 & 94.9 & 93.9 & 94.2\\ 
Zhang et al.*   & 97.3 & 92.3 & 86.8  &84.2 & 91.9 & 92.2 & 90.9 & 90.8\\ 
Artacho et al.\cite{artacho2020unipose}  & $-$ &$-$ & $-$  &$-$ & $-$ & $-$ & $-$ & 94.5 \\
Tang et al.*  & 98.3 &95.9 & 93.5  &90.7 & 95.0 & 96.6 & 95.7 & 95.1 \\
Xiao et al.*  & 98.3 &96.3 & \textbf{96.2}  &\textbf{95.1} & 96.0 & \textbf{96.7} & \textbf{95.9} & \textbf{96.4} \\
Ours(GM-SCENet) & 93.1 & \textbf{96.6} & 96.1 &94.6 & \textbf{96.4} & 95.9  & 92.7 & 95.1\\
\bottomrule
\end{tabular}
      \begin{tablenotes} 
		\item * The corresponding references are on the Supplementary H. For data preprocessing, we follow the default setting in \cite{artacho2020unipose} where we simply use the original image size as the rough scale, and the image center as the rough position of the target human to crop the image patches. 
     \end{tablenotes} 
\end{threeparttable} 
\end{center}
\end{table*}

\clearpage
\onecolumn
\section*{Supplementary E}

\begin{figure*}[htbp]
\begin{center}
\includegraphics[width=12.4cm]{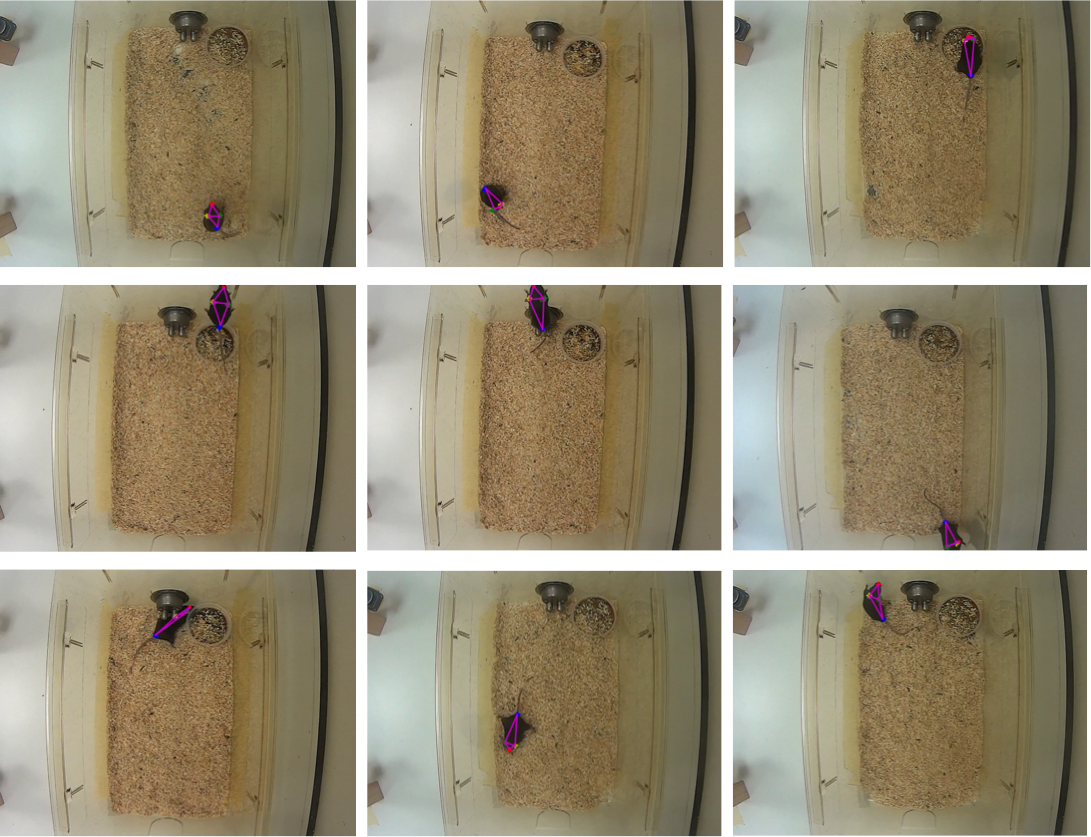}\\
(a) PDMB  \\

\includegraphics[width=12.4cm]{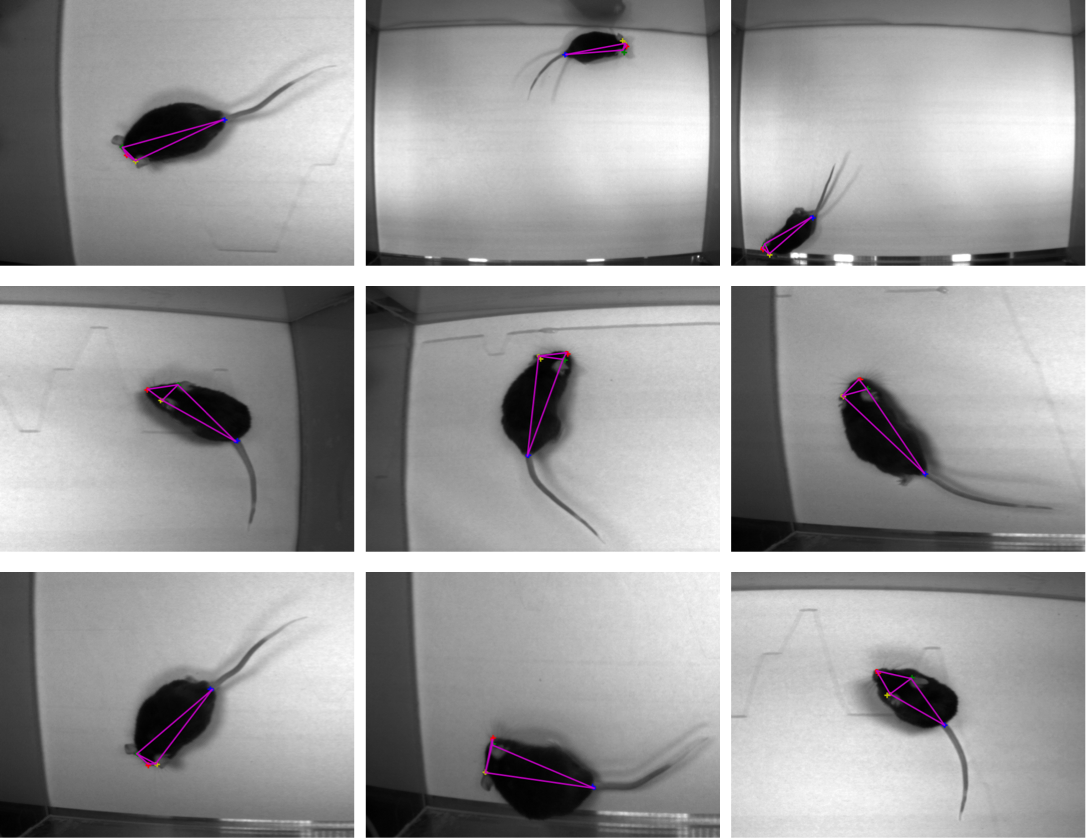}\\
(b) DeepLabCut Mouse Pose
\end{center}
\caption{Examples of the estimated mouse poses on the PDMB and DeepLabCut test sets (best viewed in electronic form with 4$\times$ zoom in). Our proposed method deals well with deformable mouse body on the PDMB dataset. The first, second and third row in (a) show occlusions, invisible keypoints and abnormal poses, respectively. On the DeeplabCut Mouse Pose dataset, our method is robust against scale variations.}
\label{fig:mousevis}
\end{figure*}

\clearpage
\onecolumn
\section*{Supplementary F}

\begin{table}[htb]
\begin{center}
\caption{Zebra, Fly and LSP Datasets used for model comparisons.}
\label{tab:datasets}
\begin{tabular}{p{2.6cm}p{3cm}p{1.6cm}p{1.6cm}p{1.6cm}p{1.6cm}}
\toprule
Name\quad & Species & Resolution & Images & Keypoints & Individuals\\
\midrule
Zebra\cite{Graving2019}  & Equus grevyi & 160*160  & 900 & 9 & Multiple\\

Fly\cite{Pereira2019}  & Drosophila melanogaster & 192*192 & 1500 & 32  & Single\\

LSP\cite{johnson2010clustered}  & Human & $ -$ & 12000 & 14  & Multiple\\
\bottomrule
\end{tabular}
\end{center}
\end{table}

\begin{figure*}[htbp]
\begin{center}
\includegraphics[width=17.4cm]{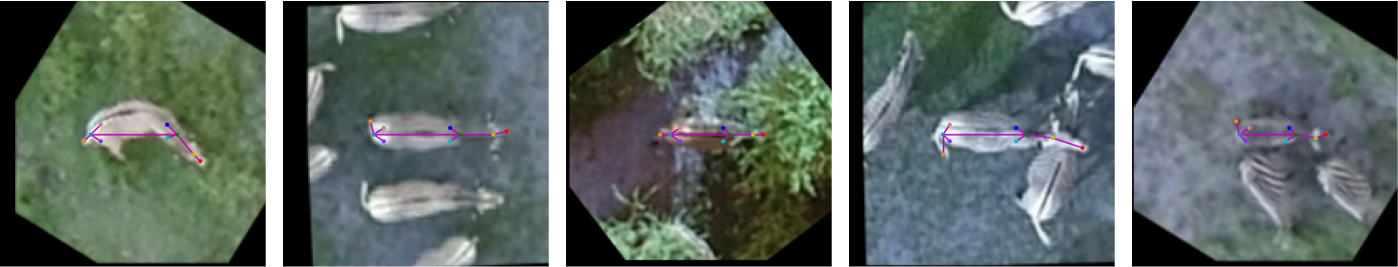}\\
(a) Zebra  \\

\includegraphics[width=17.4cm]{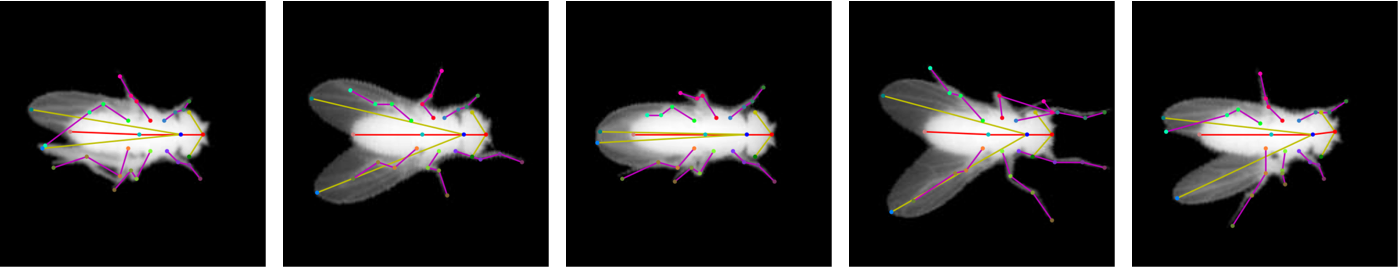}\\
(b) Fly
\end{center}
\caption{Examples of the estimated poses on the Zebra and Fly datasets. The proposed method can achieve accurate localisation on the both datasets.}
\label{fig:zebravis}
\end{figure*}

The herds of zebras are recorded in the wild. This dataset features multiple interacting individuals with highly-variable environments and lighting conditions. When multiple zebras occurs in an image, the owner of the dataset only provides the grount-truth annotation (i.e., the locations of 9 keypoints) of one of these zebras, which lies in the center of the image, as shown in Fig. \ref{fig:zebravis}(a). In our experiments, we randomly split the Zebra dataset into a training set of 720 images and a test set of 180 images. Fly dataset is recorded in a laboratory setting. Flies move freely in a backlit 100-mm-diameter circular arena covered by a 2-mm-tall clear polyethylene terephthalate glycol dome. This dataset is divided into training set (1200 images) and testing set (300 images). These datasets are freely-available from \href{https://github.com/jgraving/deepposekit-data}{https://github.com/jgraving/deepposekit-data.}

The Leeds Sports Pose (LSP) dataset is leveraged for single person pose estimation. Images for LSP are collected from Flickr for a wide range of individuals performing sports activities. The dataset includes 11000 images for training and 1000 images for testing where 14 keypoints in the entire body are labeled.

\clearpage
\onecolumn
\section*{Supplementary G}

\begin{figure*}[htbp]
\begin{center}
\includegraphics[width=16.4cm]{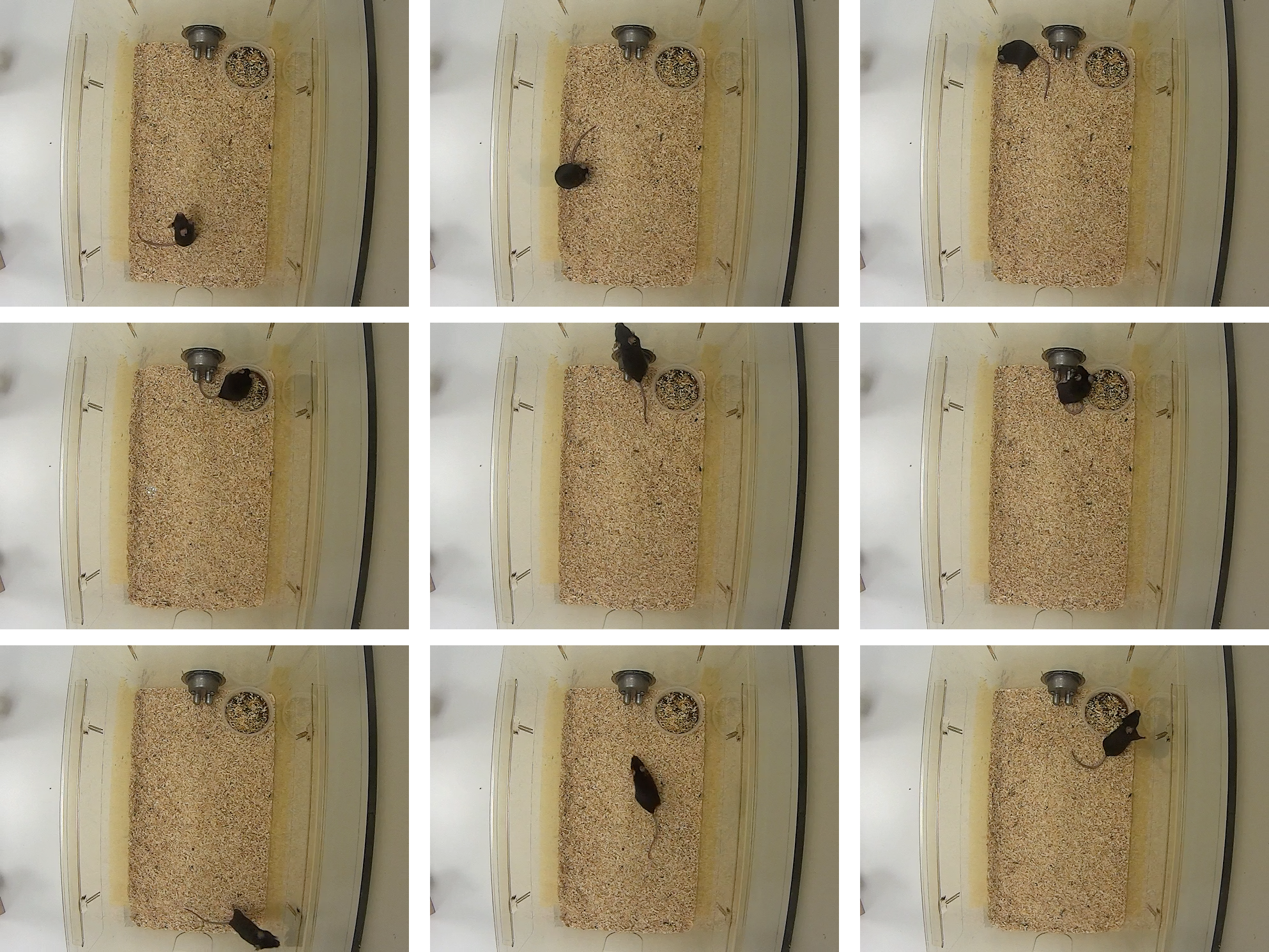}\\

\end{center}
\caption{Examples of our PDMB dataset.}
\label{fig:PDMB}
\end{figure*}

\begin{figure*}[htbp]
\begin{center}
\includegraphics[width=5.4cm]{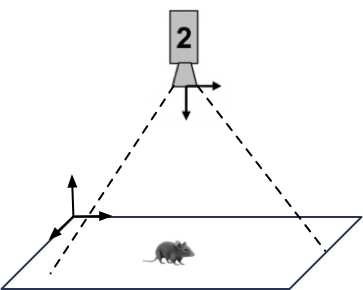}\\
\end{center}
\caption{ The location of the camera used in our mouse pose PDMB dataset. The camera is about 50cm from the bottom of the cage. The location of the camera in all experiments is fixed.}
\label{fig:camsetting}
\end{figure*}

\begin{figure*}[htbp]
\begin{center}
\includegraphics[width=16.4cm]{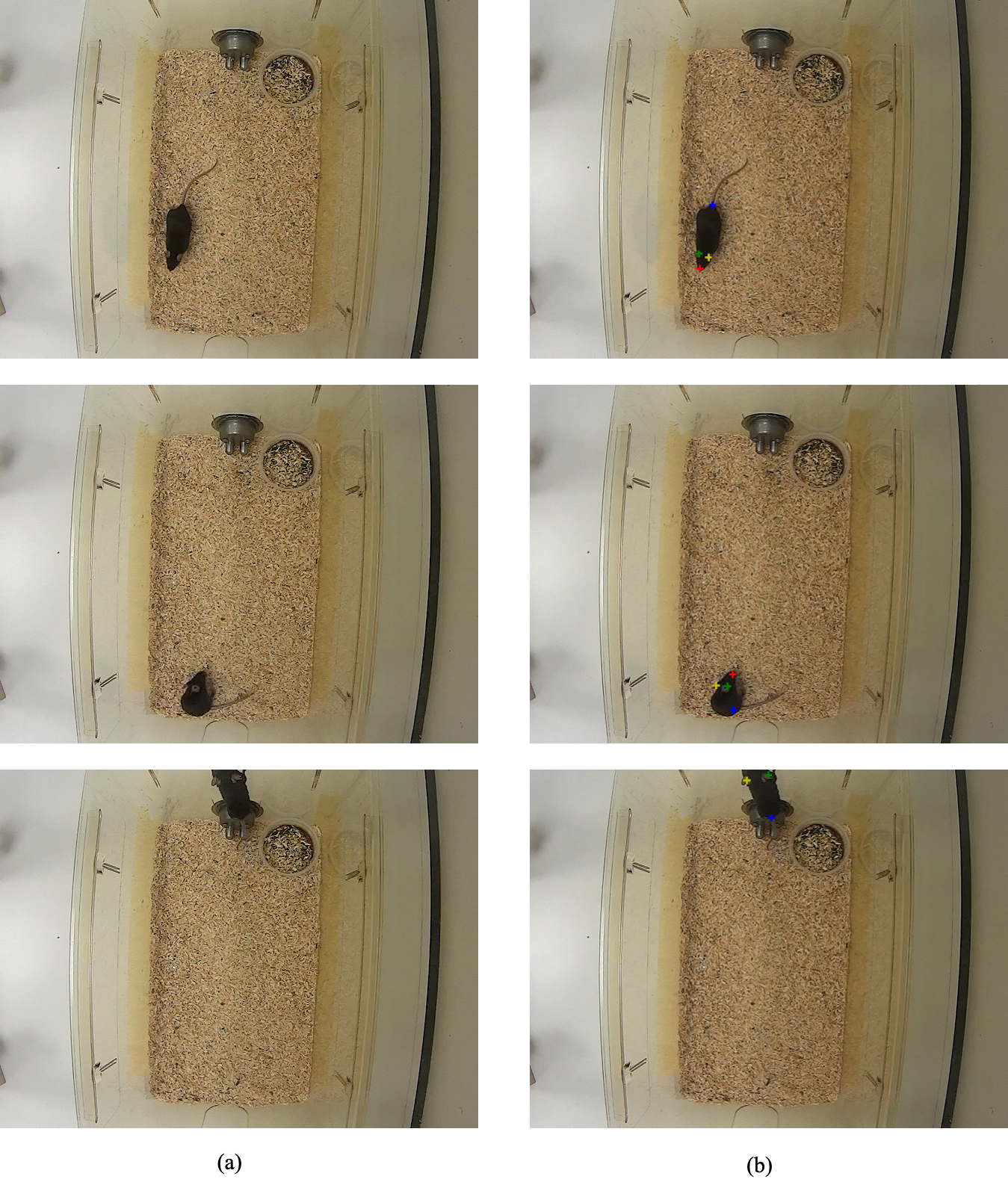}\\
\end{center}
\caption{Snapshots taken from top-view camera for different mouse poses. (a) and (b) show the original images and manual annotation results. In the first row of (b), all keypoints are visible, and the flag of the visibility of mouse parts is set to 1. The second row of (b) shows the case where a part, i.e., tail base, may be occluded. The tail base is not visible but its position is apparent given the context of the image. Our dataset provides ground truth locations for these keypoints and an additional annotation indicates their lack of visibility. The last row of (b) shows that the snout is truncated. In such case, Our dataset will not provide a ground-truth annotation. }
\label{fig:gtannotation}
\end{figure*}

 \textbf{Pose annotation}. All extracted video frames were annotated using a freeware DeepLabCut (available at \href{https://github.com/DeepLabCut/DeepLabCut}{https://github.com/DeepLabCut/DeepLabCut}). A team of six professionals were trained to annotate home-cage mouse poses. Following the experimental setting of DeepLabcut \cite{Mathis2018}, we also annotated the locations of four body parts, i.e., snout, left ear, right ear and tail base. Additionally, we annotated the visibility of mouse parts, as shown in Fig \ref{fig:gtannotation}. After data annotation, we performed secondary screening to remove ambiguous frames, leaving 9248 frames to establish our mouse pose PDMB dataset.

\clearpage
\onecolumn
\section*{Supplementary H}

\textbf{Reference}

[1] I. Lifshitz, E. Fetaya, and S. Ullman, “Human pose estimation using deep consensus voting,” in European Conference on Computer Vision. Springer, 2016, pp. 246–260.

[2] A. Bulat and G. Tzimiropoulos, “Human pose estimation via convolutional part heatmap regression,” in European Conference on Computer Vision. Springer, 2016, pp. 717–732.

[3] K. Sun, C. Lan, J. Xing, W. Zeng, D. Liu, and J. Wang, “Human pose estimation using global and local normalization,” in Proceedings of the IEEE International Conference on Computer Vision, 2017, pp. 5599– 5607.

[4] W. Yang, S. Li, W. Ouyang, H. Li, and X. Wang, “Learning feature pyramids for human pose estimation,” in proceedings of the IEEE international conference on computer vision, 2017, pp. 1281–1290.

[5] H. Zhang, H. Ouyang, S. Liu, X. Qi, X. Shen, R. Yang, and J. Jia, “Human pose estimation with spatial contextual information,” arXiv preprint arXiv:1901.01760, 2019. 

[6] Z. Cao, R. Wang, X. Wang, Z. Liu, and X. Zhu, “Improving human pose estimation with self-attention generative adversarial networks,” in 2019 IEEE International Conference on Multimedia $\&$ Expo Workshops (ICMEW). IEEE, 2019, pp. 567–572. 

[7] F. Zhang, X. Zhu, and M. Ye, “Fast human pose estimation,” in Proceedings of the IEEE/CVF Conference on Computer Vision and Pattern Recognition, 2019, pp. 3517–3526. 

[8] W. Tang, P. Yu, and Y. Wu, “Deeply learned compositional models for human pose estimation,” in Proceedings of the European conference on computer vision (ECCV), 2018, pp. 190–206. 

[9] Y. Xiao, D. Yu, X. Wang, T. Lv, Y. Fan, and L. Wu, “Spcnet: Spatial preserve and content-aware network for human pose estimation,” arXiv preprint arXiv:2004.05834, 2020.


\end{document}